\RequirePackage{fix-cm}
\documentclass[smallextended]{svjour3}       
\smartqed  
\usepackage{graphicx}

\usepackage{setspace}
\usepackage[numbers]{natbib}

\usepackage{booktabs}
\usepackage{bigstrut}
\usepackage{multirow}
\usepackage[short]{optidef}
\usepackage{enumerate}
\usepackage{subcaption}
\usepackage{rotating}
\usepackage[table,xcdraw]{xcolor}
\usepackage{method}
\usepackage{amssymb}

\usepackage{url}
\usepackage{csvsimple}
\usepackage{epstopdf}
\usepackage{float}
\usepackage{bm}
\usepackage{footmisc}
\DeclareUrlCommand\ULurl{}
\usepackage{adjustbox}
\mathchardef\mhyphen="2D 

\DeclareSymbolFont{matha}{OML}{txmi}{m}{it}
\DeclareMathSymbol{\varv}{\mathord}{matha}{118}

\usepackage{mathtools}

\DeclarePairedDelimiter\floor{\lfloor}{\rfloor}

\usepackage{threeparttable}
\usepackage{makecell}

\usepackage{algorithm}
\usepackage{algpseudocode}
\usepackage{etoolbox}
\usepackage{tikz}
\usetikzlibrary{tikzmark}
\usetikzlibrary{calc}

\errorcontextlines\maxdimen

\newcommand{\ALGtikzmarkcolor}{black}
\newcommand{\ALGtikzmarkextraindent}{4pt}
\newcommand{\ALGtikzmarkverticaloffsetstart}{-.5ex}
\newcommand{\ALGtikzmarkverticaloffsetend}{-.5ex}
\makeatletter
\newcounter{ALG@tikzmark@tempcnta}

\newcommand\ALG@tikzmark@start{%
	\global\let\ALG@tikzmark@last\ALG@tikzmark@starttext%
	\expandafter\edef\csname ALG@tikzmark@\theALG@nested\endcsname{\theALG@tikzmark@tempcnta}%
	\tikzmark{ALG@tikzmark@start@\csname ALG@tikzmark@\theALG@nested\endcsname}%
	\addtocounter{ALG@tikzmark@tempcnta}{1}%
}

\def\ALG@tikzmark@starttext{start}
\newcommand\ALG@tikzmark@end{%
	\ifx\ALG@tikzmark@last\ALG@tikzmark@starttext
	\else
	\tikzmark{ALG@tikzmark@end@\csname ALG@tikzmark@\theALG@nested\endcsname}%
	\tikz[overlay,remember picture] \draw[\ALGtikzmarkcolor] let \p{S}=($(pic cs:ALG@tikzmark@start@\csname ALG@tikzmark@\theALG@nested\endcsname)+(\ALGtikzmarkextraindent,\ALGtikzmarkverticaloffsetstart)$), \p{E}=($(pic cs:ALG@tikzmark@end@\csname ALG@tikzmark@\theALG@nested\endcsname)+(\ALGtikzmarkextraindent,\ALGtikzmarkverticaloffsetend)$) in (\x{S},\y{S})--(\x{S},\y{E});%
	\fi
	\gdef\ALG@tikzmark@last{end}%
}

\apptocmd{\ALG@beginblock}{\ALG@tikzmark@start}{}{\errmessage{failed to patch}}
\pretocmd{\ALG@endblock}{\ALG@tikzmark@end}{}{\errmessage{failed to patch}}
\makeatother

\usepackage{array}

\begin{document}

\title{Minimum Variance Embedded Auto-associative Kernel Extreme Learning Machine for One-class Classification}

\titlerunning{Minimum Variance Embedded Auto-associative KELM for OCC}        

\author{Pratik K. Mishra  	\and
		Chandan Gautam  	\and
		Aruna Tiwari
}


\institute{Pratik K. Mishra \at
	Indian Institute of Technology Indore, India \\
	\email{ms1804101003@iiti.ac.in, mpratik995@gmail.com} 
	\and
	Chandan Gautam \at
	Indian Institute of Technology Indore, India \\
	\email{phd1501101001@iiti.ac.in, chandangautam31@gmail.com} 
	\and
	Aruna Tiwari \at
	Indian Institute of Technology Indore, India \\
	\email{artiwari@iiti.ac.in}
}

\date{Received: date / Accepted: date}

\maketitle

\begin{abstract}
One-class classification (OCC) needs samples from only a single class to train the classifier. Recently, an auto-associative kernel extreme learning machine was developed for the OCC task. This paper introduces a novel extension of this classifier by embedding minimum variance information within its architecture and is referred to as \textit{VAAKELM}. The minimum variance embedding forces the network output weights to focus in regions of low variance and reduces the intra-class variance. This leads to a better separation of target samples and outliers, resulting in an improvement in the generalization performance of the classifier. The proposed classifier follows a reconstruction-based approach to OCC and minimizes the reconstruction error by using the kernel extreme learning machine as the base classifier. It uses the deviation in reconstruction error to identify the outliers. We perform experiments on 15 small-size and 10 medium-size one-class benchmark datasets to demonstrate the efficiency of the proposed classifier. We compare the results with 13 existing one-class classifiers by considering the mean F$_1$ score as the comparison metric. The experimental results show that \textit{VAAKELM} consistently performs better than the existing classifiers, making it a viable alternative for the OCC task.

\keywords{Minimum Variance Embedding \and Kernel Extreme Learning Machine \and Reconstruction-based \and One-Class Classification}
\end{abstract}

\section{Introduction}
\label{intro}
One-class classification (OCC) has been an extensive area of research in recent years. It has been employed in various real-world scenarios for anomaly or novelty detection \cite{GRINBLAT20137242,pimentel2014review,CABRAL20147182,BELLINGER2018,sohrab2018subspace,8692763}. In many real-world scenarios, acquiring samples of the class of interest (i.e., normal or positive class) is relatively easier in comparison to acquiring the negative class samples, which are very rare or expensive to collect. For example, building a classifier to identify if a person is healthy or not. In such a scenario, collecting data for the healthy cases is quite easy as the characteristics of all the healthy persons are quite similar. However, it is quite difficult to collect information about all the unhealthy cases in the world. In such a case, a classifier needs to be trained using healthy class samples only, for which a one-class classifier is the best-suited solution as it needs samples of only a single class. OCC has been employed frequently in the past for many real-world applications, like fault detection \cite{shin2005one}, document classification \cite{manevitz2007one}, authorship verification \cite{koppel2004authorship}, video surveillance \cite{diehl2002real,markou2006neural}, and intrusion detection \cite{Fan2004}.

Different one-class classifiers have been proposed in the past \cite{MARKOU20032481,MARKOU20032499,khan2014one,pimentel2014review}. These classifiers can be broadly classified into three categories \cite{tax2002one}, namely, (i) density-based  (ii) boundary-based (iii) reconstruction-based. In the density-based classifiers, we estimate the density of the training data and then apply a certain threshold. The sample size must be sufficiently large to give a good generalization performance. Some of the early works in the area of density-based classifiers include parzen density estimation \cite{parzen1962estimation}. Here, the training data is used to estimate the probability density of the target. The samples whose estimated probability is lesser than the threshold are classified as outliers. In the boundary-based classifiers, a closed boundary around the target set is optimized. Boundary-based classifiers need less number of samples in comparison to density-based classifiers. However, as they mostly rely on the distance between samples, they tend to be sensitive to the scaling of features. Boundary-based classifiers can further be categorized into non-kernel and kernel-based classifiers. Some of the non-kernel based classifiers are k-centers \cite{ypma1998support} and k-nearest neighbors \cite{knorr2000distance}. Support vector machine (SVM) was used as a base classifier to develop the kernel-based classifiers. In one-class support vector machine (OCSVM) \cite{scholkopf2001estimating}, a hyperplane is used to separate the region that contains no data. The hyperplane is placed at a maximum distance from the origin. Instead of a hyperplane, Tax and Duin \cite{tax2004support} used a hypersphere to include maximum training data with minimum radius, referred to as support vector data description (SVDD). The reconstruction-based classifiers are used to obtain a more compact representation of the target data by making assumptions about the target distribution. The reconstruction error is used to distinguish normal class from outliers based on the idea that the outlier objects do not satisfy the assumptions about target distribution, and their reconstruction error should be high. In reconstruction-based classifiers, prior knowledge of the data is required. Various methods have been taken as base classifiers to develop different reconstruction-based one-class classifiers. In the k-means clustering-based one-class classifier \cite{jiang2001two}, it is assumed that the data is clustered and can be characterized by a few prototype objects. Bishop et al. \cite{bishop1995neural} developed a Principal component analysis (PCA) based one-class classifier. Some of the neural network-based approaches are Autoencoder or Multi-layer Perceptron (MLP) \cite{japkowicz1995novelty} and diabolo networks \cite{Hertz:1991:ITN:574634,baldi1989neural}. They are trained to reproduce the input pattern at the output layer. However, they inherit the same problems as in the conventional application of neural networks, requiring a predefined number of layers and neurons, learning rate, and the stopping criterion from the user. One Class Random Forests (OCRF) \cite{desir2013one} is an ensemble-based one-class classifier that works by combining several weak classifiers known to be accurate. It efficiently generates outliers by subsampling the training dataset. Over the past few years, Extreme Learning Machines (ELMs) have been applied for OCC \cite{leng2015one,mygdalis2016one} due to their non-iterative nature.

ELMs \cite{huang2004extreme,huang2006extreme} were originally proposed for single-hidden layer feed-forward neural networks (SLFNs). The purpose of ELM was to tackle the slow learning speed of gradient-based algorithms used in traditional neural networks. The essence of ELM is that the hidden layer parameters need not be tuned, as the hidden layer weights and biases are randomly assigned. The training involves analytically determining the output weights by reducing the training error, which in turn provides a good generalization and faster performance as it does not require tuning of parameters. 
Leng et al. \cite{leng2015one} proposed a one-class kernel ELM-based (\textit{OCKELM}) one-class classifier where the hidden layers were replaced by a kernel. It followed a boundary-based approach to OCC and used the kernel extreme learning machine (KELM) as the base classifier. Mygdalis et al. \cite{mygdalis2016one} proposed a minimum-variance embedded KELM-based (\textit{VOCKELM}) one class classifier that minimizes the training error and reduces the variance of the class data. Gautam et al. \cite{gautam2017construction} proposed an auto-associative KELM-based (\textit{AAKELM}) one-class classifier, where they leveraged the reconstruction error from an autoencoder for OCC. In this paper, we propose a novel extension of the existing \textit{AAKELM} classifier by embedding minimum variance information within its architecture. The proposed classifier performs better than the existing classifiers, \textit{OCKELM} and \textit{VOCKELM}, as it uses a kernel autoencoder to reconstruct the essential features from the input data. The minimum variance embedding reduces the intra-class variance and aids in better separation of target class samples from outliers, thus showing an improvement in performance over the existing \textit{AAKELM} classifier.

In this paper, we propose a minimum variance embedded auto-associative kernel extreme learning machine (\textit{VAAKELM}) for OCC. \textit{VAAKELM} incorporates minimum-variance embedding to minimize the data dispersion and enable better separation of outliers. It also employs a kernel autoencoder to learn the essential information from the input data. \textit{VAAKELM} reduces the intra-class variance and reconstruction error, simultaneously. It uses the main idea behind the reconstruction-based methods; that is, the outlier objects do not satisfy the assumptions about target distribution, and their reconstruction error should be high. We evaluate the performance of \textit{VAAKELM} on 15 small-size and 10 medium-size one-class benchmark datasets and compare its performance with 13 existing one-class classifiers using the F$_1$ score as the performance metric.

The rest of the paper is organized as follows. Section \ref{sec_preliminary} discusses the prerequisite KELM-based one-class classifiers. Section \ref{sec_proposed} describes our proposed classifier. Section \ref{sec_exp} discusses the performance metrics and the experimental results. Finally, we conclude our paper in Section \ref{sec_conclusion}.

\section{Preliminaries} \label{sec_preliminary}
The proposed work is a KELM-based one-class classifier. Hence, we discuss the existing KELM-based one-class classifiers in this section, that should enable a better understanding of our proposed work.

\subsection{OCKELM}
One-class kernel-based ELM \cite{leng2015one} is a boundary-based classifier. Taking the training input as,  $\textbf{X}=\{\boldsymbol{x}_i \: | \: \boldsymbol{x}_i \in R^d, i=1,2,...,\mathcal{N}\}$, and $\textbf{r}=[r,...,r]^T \in R^\mathcal{N}$. $r$ is a real number, referred to as the target class. The training involves minimizing output weight ($\boldsymbol{\beta}$) using the following optimization criterion,
\begin{align} \label{OCKELM_min}
& \min_{\boldsymbol{\beta}, e_i} \: \frac{1}{2} \: ||\boldsymbol{\beta}||_2^2 + \frac{1}{2}C \: \sum_{i=1}^{\mathcal{N}} ||e_i||_2^2 \\
& s.t. \: \: \boldsymbol{\beta}^T \boldsymbol{\mathnormal{h}}(\boldsymbol{x}_i) = r - e_i, \: i = 1,2,...,\mathcal{N}, \nonumber
\end{align}
where, $e_i$ is the training error, and $\boldsymbol{\mathnormal{h}}(\boldsymbol{x}_i)$ is the hidden layer output for a sample $\boldsymbol{x}_i$. $C$ is the regularization parameter. Solving above equation, the output weight is derived as,
\begin{equation} \label{OCKELM_beta}
\boldsymbol{\beta} = \mathcal{H}^T \left( \frac{1}{C} \mathbb{I} + \mathcal{H}\mathcal{H}^T \right)^{-1} \textbf{r},
\end{equation}
where, $\mathcal{H}$ denotes the hidden layer output matrix and $\mathbb{I}$ is an identity matrix. The network output is then expressed as, $\boldsymbol{\widehat{O}} = \boldsymbol{\mathnormal{h}}(\boldsymbol{x}) \: \boldsymbol{\beta}$. Making use of Mercer's conditions, kernel matrix $\boldsymbol{\mathit{\Omega}}$ is defined as $\boldsymbol{\mathit{\Omega}}=\mathcal{H}\mathcal{H}^T$,  $\mathit{\Omega}_{j,k}=\boldsymbol{\mathnormal{h}}(\boldsymbol{x}_j)\boldsymbol{\mathnormal{h}}(\boldsymbol{x}_k)=K(\boldsymbol{x}_j,\boldsymbol{x}_k), \: j,k = 1,...,\mathcal{N}$. $K$ is a kernel function. The output weight $\boldsymbol{\beta}$ is then expressed as,
\begin{equation} \label{OCKELM_betaK}
\boldsymbol{\beta} = \left( \frac{1}{C} \mathbb{I} + \boldsymbol{\mathit{\Omega}} \right)^{-1} \textbf{r}.
\end{equation}
The network output $\boldsymbol{\widehat{O}}$ is further expressed as,
\begin{equation} \label{OCKELM_outK}
\boldsymbol{\widehat{O}} = \begin{bmatrix}
K(\boldsymbol{x},\boldsymbol{x}_1) \\
\vdots\\
K(\boldsymbol{x},\boldsymbol{x}_\mathcal{N})
\end{bmatrix}^T  \left( \frac{1}{C} \mathbb{I} + \boldsymbol{\mathit{\Omega}} \right)^{-1} \textbf{r}.
\end{equation}
The deviation of the network outputs is then determined as, $\boldsymbol{s} = \left| \boldsymbol{\widehat{O}} - \textbf{r} \right|$. Defining $\boldsymbol{s}_{dec}$ as vector $\boldsymbol{s}$ sorted in decreasing order, the threshold ($\theta$) is calculated as,
\begin{equation} \label{OCKELM_theta}
\theta=\boldsymbol{s}_{dec}(\floor{\delta * \mathcal{N}}),    
\end{equation}
where, $\delta$ is the fraction of dismissal. 

\noindent For a test sample $\boldsymbol{x}_t$, the network output $\widehat{O}_t$ is determined as,
\begin{equation} \label{OCKELM_outK_test}
\widehat{O}_t = \begin{bmatrix}
K(\boldsymbol{x}_t,\boldsymbol{x}_1) \\
\vdots\\
K(\boldsymbol{x}_t,\boldsymbol{x}_\mathcal{N})
\end{bmatrix}^T  \boldsymbol{\beta}.
\end{equation}
The deviation $s_t$ is determined as $s_t = \left| \widehat{O}_t - r \right|$. Finally, the classification follows the decision rule,
\begin{align} \label{OCKELM_decision}
sign(\theta-s_t) &= 1, \: \: \: \boldsymbol{x}_t \: belongs \: to \: target \: class. \\
& \: -1, \: \: \: \boldsymbol{x}_t \: belongs \: to \: outlier \: class. \nonumber
\end{align}


\subsection{AAKELM} \label{sec_AAKELM}
In auto-associative kernel-based ELM \cite{gautam2017construction}, the data at the input layer is used for reconstruction at the output layer using kernelized feature mapping. Being auto-associative in nature, the input and output layer is made of an equal number of nodes. $\textbf{X}=\{\boldsymbol{x}_i \: | \: \boldsymbol{x}_i \in \mathbb{R}^d, i=1,2,...,\mathcal{N}\}$ is taken as the training input. The training involves calculating optimum output weight, $\boldsymbol{\beta}$, by solving the following optimization problem,
\begin{align} \label{AAKELM_min}
& \min_{\boldsymbol{\beta},\textbf{e}_i} \: \frac{1}{2} \: ||\boldsymbol{\beta}||_F^2 + \frac{1}{2}C \: \sum_{i=1}^{\mathcal{N}} ||\textbf{e}_i||_2^2 \\
& s.t \: \: \boldsymbol{\beta}^T \boldsymbol{\mathnormal{h}}(\boldsymbol{x}_i) = \boldsymbol{x}_i - \textbf{e}_i, \: i = 1,2,...,\mathcal{N}, \nonumber
\end{align}
where, $\textbf{e}_i$ is the reconstruction error, and $\boldsymbol{\mathnormal{h}}(\boldsymbol{x}_i)$ is the hidden layer output for a sample $\boldsymbol{x}_i$. $C$ acts as the trade-off between minimizing the output weight norm and the reconstruction error. $|| \: . \: ||_F$ refers to frobenius norm. Solving equation (\ref{AAKELM_min}), the output weight is derived as,
\begin{equation} \label{AAKELM_beta}
\boldsymbol{\beta} = \mathcal{H}^T \left( \frac{1}{C} \mathbb{I} + \mathcal{H}\mathcal{H}^T \right)^{-1} \textbf{X}^T,
\end{equation}
where, $\mathcal{H}$ denotes the hidden layer output matrix, and $\mathbb{I}$ is an identity matrix. The network output is derived as, $\boldsymbol{\widehat{O}} =  \boldsymbol{\mathnormal{h}}(\boldsymbol{x}) \boldsymbol{\beta}$. 
Defining $\boldsymbol{\mathit{\Omega}}$ as a kernel matrix with kernel $K$ as $\boldsymbol{\mathit{\Omega}}=\mathcal{H}\mathcal{H}^T$,  $\mathit{\Omega}_{j,k}=\boldsymbol{\mathnormal{h}}(\boldsymbol{x}_j)\boldsymbol{\mathnormal{h}}(\boldsymbol{x}_k)=K(\boldsymbol{x}_j,\boldsymbol{x}_k), \: j,k = 1,...,\mathcal{N}$, the output weight is rewritten as,
\begin{equation} \label{AAKELM_betaK}
\boldsymbol{\beta} = \left( \frac{1}{C} \mathbb{I} + \boldsymbol{\mathit{\Omega}} \right)^{-1} \textbf{X}^T.
\end{equation}
The kernelized network output of \textit{AAKELM} is then expressed as,
\begin{equation} \label{AAKELM_outK}
\boldsymbol{\widehat{O}} = \begin{bmatrix}
K(\boldsymbol{x},\boldsymbol{x}_1) \\
\vdots\\
K(\boldsymbol{x},\boldsymbol{x}_\mathcal{N})
\end{bmatrix}^T  \left( \frac{1}{C} \mathbb{I} + \boldsymbol{\mathit{\Omega}} \right)^{-1} \textbf{X}^T.
\end{equation}
\noindent
Further, we calculate the loss $\boldsymbol{s}$ as,
\begin{equation}
s_i =\sum_{j=1}^{d}(\widehat{O}_{ij} - x_{ij})^2, \text{  }i=1,2,...,\mathcal{N}.
\end{equation}
$\boldsymbol{s}$ is then sorted in decreasing order and is denoted as, $\boldsymbol{s}_{dec}$. The threshold ($\theta$) is then calculated as,
\begin{equation} \label{AAKELM_thresh}
\theta=\boldsymbol{s}_{dec}(\floor{\delta * \mathcal{N}}),    
\end{equation}
where, $\delta$ is the fraction of dismissal. For test sample $\boldsymbol{x}_t$, the network output $\widehat{O}_t$ is determined as,
\begin{equation} \label{AAKELM_outK_test}
\widehat{O}_t = \begin{bmatrix}
K(\boldsymbol{x}_t,\boldsymbol{x}_1) \\
\vdots\\
K(\boldsymbol{x}_t,\boldsymbol{x}_\mathcal{N})
\end{bmatrix}^T  \boldsymbol{\beta}.
\end{equation}
$s_t$ is then calculated as, $s_t=\sum_{j=1}^{d}\left(\widehat{O}_{tj} - x_{tj}\right)^2$. Finally, classification is done using following rule,
\begin{align} \label{AAKELM_decision}
sign(\theta-s_t) &= 1, \: \: \: \boldsymbol{x}_t \: belongs \: to \: target \: class, \\
& \: -1, \: \: \: \boldsymbol{x}_t \: belongs \: to \: outlier \: class. \nonumber
\end{align}

\noindent


\subsection{VOCKELM}
Minimum Variance One-Class KELM \cite{mygdalis2016one} improves the generalization performance of \textit{OCKELM} by reducing the intra-class variance. $\textbf{X}=\{\boldsymbol{x}_i \: | \: \boldsymbol{x}_i \in \mathbb{R}^d, i=1,2,...,\mathcal{N}\}$ is taken as the training input, and $r$ is the target class. The training involves minimizing output weight ($\boldsymbol{\beta}$) using following optimization criterion,
\begin{align} \label{VOCKELM_min}
& \min_{\boldsymbol{\beta},e_i} \frac{1}{2} \left \Vert \boldsymbol{\beta}^T (\textbf{V}_C+ \lambda \mathbb{I}) \: \boldsymbol{\beta} \right \Vert_2^2 + \frac{C}{2} \sum_{i=1}^{\mathcal{N}} ||e_i||_2^2 \\
& s.t. \: \: \boldsymbol{\beta}^T \boldsymbol{\mathnormal{h}}(\boldsymbol{x}_i) = r - e_i, \: i = 1,2,...,\mathcal{N}, \nonumber
\end{align}
where, $e_i$ is the training error, and $\boldsymbol{\mathnormal{h}}(\boldsymbol{x}_i)$ is the hidden layer output for a sample $\boldsymbol{x}_i$. $\boldsymbol{\beta}$ is the output weight, and C acts as the trade-off between the norm of output weight and the training error. $\mathbb{I}$ is an identity matrix. $\lambda$ is a regularization parameter. The class variance ($\textbf{V}_C$) is expressed as,
\begin{align} \label{VOCKELM_CVdecomp}
\textbf{V}_C &= \frac{1}{\mathcal{N}} \: \sum_{i=1}^{\mathcal{N}} \: \left(\boldsymbol{\mathnormal{h}}(\boldsymbol{x}_i) - \overline{\mathcal{H}}\right) \left(\boldsymbol{\mathnormal{h}}(\boldsymbol{x}_i) - \overline{\mathcal{H}}\right)^T \nonumber \\
&= \frac{1}{\mathcal{N}} \: \mathcal{H} \left( \mathbb{I} - \frac{1}{\mathcal{N}} \textbf{aa}^T \right) \mathcal{H}^T, \nonumber \\
&= \mathcal{H} \: \boldsymbol{\mathcal{M}} \: \mathcal{H}^T,
\end{align}
where, \textbf{a} is a vector of ones and $\mathcal{H}$ denotes the hidden layer output matrix. $\overline{\mathcal{H}}= \frac{1}{\mathcal{N}} \sum_{i=1}^{\mathcal{N}} \boldsymbol{\mathnormal{h}}(\boldsymbol{x}_i)$ is the mean of the hidden layer outputs of the network, and $\boldsymbol{\mathcal{M}}$ represents any Laplacian matrix. 
The intra-class variance ($\textbf{V}_S$) is expressed as,
\begin{equation}  \label{VOCKELM_SV}
\textbf{V}_S = \sum_{i=1}^{\mathcal{N}} \sum_{p=1}^{P} \frac{\mathcal{N}_p}{\mathcal{N}} \gamma_i^p (\boldsymbol{\mathnormal{h}}(\boldsymbol{x}_i) - \overline{\mathcal{H}}) (\boldsymbol{\mathnormal{h}}(\boldsymbol{x}_i) - \overline{\mathcal{H}})^T,
\end{equation}
where, $\mathcal{N}_p$ denotes the number of samples that belongs to the cluster $p$, and $\gamma_i^p$ denotes if $\boldsymbol{x}_i$ belongs to cluster $p$ or not. The intra-class variance can further be expressed in a similar way as $\textbf{V}_C$ in equation (\ref{VOCKELM_CVdecomp}).

\noindent Replacing $\textbf{V}_C$ in equation (\ref{VOCKELM_min}) with the expression in equation (\ref{VOCKELM_CVdecomp}), the output weight $\boldsymbol{\beta}$ is derived as,
\begin{equation} \label{VOCKELM_beta}
\boldsymbol{\beta} = \mathcal{H}^T \left(\mathcal{H}\mathcal{H}^T + \frac{1}{C} \boldsymbol{\mathcal{M}}\mathcal{H}\mathcal{H}^T + \frac{\lambda}{C} \mathbb{I} \right)^{-1} \textbf{r}.
\end{equation}
The network output is expressed as $\boldsymbol{\widehat{O}} = \boldsymbol{\mathnormal{h}}(\boldsymbol{x}) \boldsymbol{\beta}$. Kernel matrix $\boldsymbol{\mathit{\Omega}}$ is defined as $\boldsymbol{\mathit{\Omega}}=\mathcal{H}\mathcal{H}^T$,  $\mathit{\Omega}_{j,k}=\boldsymbol{\mathnormal{h}}(\boldsymbol{x}_j)\boldsymbol{\mathnormal{h}}(\boldsymbol{x}_k)=K(\boldsymbol{x}_j,\boldsymbol{x}_k), \: j,k = 1,...,\mathcal{N}$. $K$ is a kernel function. The output weight is derived as,
\begin{equation} \label{VOCKELM_betaK}
\boldsymbol{\beta} = \left(\boldsymbol{\mathit{\Omega}} + \frac{1}{C} \boldsymbol{\mathcal{M}}\boldsymbol{\mathit{\Omega}} + \frac{\lambda}{C} \mathbb{I} \right)^{-1} \textbf{r}.
\end{equation}
Further, the network output, $\boldsymbol{\widehat{O}}$, is then calculated as,
\begin{equation} \label{VOCKELM_outK}
\boldsymbol{\widehat{O}} = \begin{bmatrix}
K(\boldsymbol{x},\boldsymbol{x}_1) \\
\vdots\\
K(\boldsymbol{x},\boldsymbol{x}_\mathcal{N})
\end{bmatrix}^T  \left(\boldsymbol{\mathit{\Omega}} + \frac{1}{C} \boldsymbol{\mathcal{M}}\boldsymbol{\mathit{\Omega}} + \frac{\lambda}{C} \mathbb{I} \right)^{-1} \textbf{r}.
\end{equation}
The threshold ($\theta$) is calculated during training as,
\begin{equation} \label{VOCKELM_thresh}
\theta = \delta \: \overline{O}, 
\end{equation}
where, $\overline{O}$ is the mean network output of training samples and $\delta$ is the fraction of dismissal.

\noindent At the time of testing, the network output for the test sample $\boldsymbol{x}_t$ is determined as,
\begin{equation} \label{VOCKELM_outK_test}
\widehat{O}_t = \begin{bmatrix}
K(\boldsymbol{x}_t,\boldsymbol{x}_1) \\
\vdots\\
K(\boldsymbol{x}_t,\boldsymbol{x}_\mathcal{N})
\end{bmatrix}^T  \boldsymbol{\beta}.
\end{equation}
Finally, the classification follows the decision rule,
\begin{equation} \label{VOCKELM_decision}
\left( \widehat{O}_t - r \right)^2 \leq \theta.
\end{equation}


\section{The Proposed One-class Classifier} \label{sec_proposed}
This section discusses the proposed classifier: minimum \textbf{v}ariance embedded \textbf{a}uto-\textbf{a}ssociative \textbf{k}ernel \textbf{e}xtreme \textbf{l}earning \textbf{m}achine for OCC (\textit{VAAKELM}). \textit{VAAKELM} is a novel extension to an existing reconstruction-based classifier for OCC \cite{gautam2017construction}. \textit{VAAKELM} is a single-layer-based one-class classifier with the auto-associative KELM as the base classifier. The minimum variance embedding minimizes the intra-class variance and forces the network output weights to focus in low-variance regions. 
Further, it uses kernel autoencoders to learn the essential information of the input data. This improves the generalization performance of the model, resulting in better OCC. \textit{VAAKELM} follows a reconstruction-based approach to OCC, hence it uses the deviation in reconstruction error between the input data and the network output to empirically determine a threshold that helps to identify the target class samples. Further, we discuss the mathematical formulation of the proposed classifier.

Taking the training input as, $\textbf{X}=\{\boldsymbol{x}_i \: | \: \boldsymbol{x}_i \in \mathbb{R}^d, i=1,2,...,\mathcal{N}\}$, the variance of the output is expressed as,
\begin{align} 
\textbf{V} &= \frac{1}{\mathcal{N}} \: \sum_{i=1}^{\mathcal{N}} \: \left( \widehat{O}_i - \overline{O} \right) \: \left( \widehat{O}_i - \overline{O} \right)^T,  \nonumber \\
&= \frac{1}{\mathcal{N}} \: \sum_{i=1}^{\mathcal{N}} \: \left(\left(\boldsymbol{\beta}\right)^T \boldsymbol{\mathnormal{h}}(\boldsymbol{x}_i) - \left(\boldsymbol{\beta}\right)^T \overline{\mathcal{H}}\right) \: \left(\left(\boldsymbol{\beta}\right)^T \boldsymbol{\mathnormal{h}}(\boldsymbol{x}_i) - \left(\boldsymbol{\beta}\right)^T \overline{\mathcal{H}}\right)^T,  \nonumber \\
&= \left( \boldsymbol{\beta} \right)^T \left( \frac{1}{\mathcal{N}} \: \sum_{i=1}^{\mathcal{N}} \: \left(\boldsymbol{\mathnormal{h}}(\boldsymbol{x}_i) - \overline{\mathcal{H}}\right) \left(\boldsymbol{\mathnormal{h}}(\boldsymbol{x}_i) - \overline{\mathcal{H}}\right)^T \right) \boldsymbol{\beta}, \nonumber \\
&= \left( \boldsymbol{\beta} \right)^T \textbf{V}_C \: \boldsymbol{\beta},
\end{align}
where, $\widehat{O}_i$ is the network output and $\boldsymbol{\mathnormal{h}}(\boldsymbol{x}_i)$ is the hidden layer output for a training sample $\boldsymbol{x}_i$. $\boldsymbol{\beta}$ is the output weight, and $\overline{O}=\frac{1}{\mathcal{N}} \sum_{i=1}^{\mathcal{N}} \widehat{O}_i$ is the mean network output for all training samples. 
$\overline{\mathcal{H}}= \frac{1}{\mathcal{N}} \sum_{i=1}^{\mathcal{N}} \boldsymbol{\mathnormal{h}}(\boldsymbol{x}_i)$ is the mean of the hidden layer outputs of the network, and $\textbf{V}_C$ is the class variance.

Either class or intra-class variance is minimized to perform minimum-variance embedding in \textit{VAAKELM}. Reproducing the class variance ($\textbf{V}_C$) from equation (\ref{VOCKELM_CVdecomp}),
\begin{align}  \label{model_CVdecomp}
\textbf{V}_C &= \frac{1}{\mathcal{N}} \sum_{i=1}^{\mathcal{N}} (\boldsymbol{\mathnormal{h}}(\boldsymbol{x}_i) - \overline{\mathcal{H}})(\boldsymbol{\mathnormal{h}}(\boldsymbol{x}_i) - \overline{\mathcal{H}})^T \nonumber \\
&= \frac{1}{\mathcal{N}} \mathcal{H} \left( \mathbb{I}-\frac{1}{\mathcal{N}}\textbf{aa}^T \right) \left( \mathcal{H} \right)^T \nonumber \\
&= \mathcal{H} \boldsymbol{\mathcal{M}} \left( \mathcal{H} \right)^T,
\end{align}
where, $\mathcal{H}$ denotes the hidden layer output matrix, $\mathbb{I}$ is an identity matrix, \textbf{a} is a vector of ones, and $\boldsymbol{\mathcal{M}}$ represents any Laplacian matrix. Similarly, the intra-class variance ($\textbf{V}_S$) can be reproduced from equation \ref{VOCKELM_SV}.
\begin{equation}  \label{model_SV}
\textbf{V}_S = \sum_{i=1}^{\mathcal{N}} \sum_{p=1}^{P} \frac{\mathcal{N}_p}{\mathcal{N}} \gamma_i^p (\boldsymbol{\mathnormal{h}}(\boldsymbol{x}_i) - \overline{\mathcal{H}}) (\boldsymbol{\mathnormal{h}}(\boldsymbol{x}_i) - \overline{\mathcal{H}})^T,
\end{equation}
where, $\mathcal{N}_p$ denotes the number of samples that belongs to the cluster $p$, and $\gamma_i^p$ denotes if $\boldsymbol{x}_i$ belongs to cluster $p$ or not. The data is grouped into subclasses using a clustering method like k-means. The intra-class variance can further be represented in a similar way as $\textbf{V}_C$ in equation (\ref{model_CVdecomp}). 

\textit{VAAKELM} minimizes the intra-class variance and the reconstruction error, simultaneously, using the proposed optimization criterion,
\begin{align} \label{model_min}
& \min_{\boldsymbol{\beta},\textbf{e}_i} \frac{1}{2} \left| \left| \left(\boldsymbol{\beta}\right)^T (\textbf{V}_C+ \lambda \mathbb{I}) \: \boldsymbol{\beta} \right| \right|_F^2 + \frac{C}{2} \: \sum_{i=1}^{\mathcal{N}} \left \Vert \textbf{e}_i \right \Vert_2^2\\
& s.t. \: \: \left( \boldsymbol{\beta} \right)^T \boldsymbol{\mathnormal{h}} \left(\boldsymbol{x}_i \right) = \boldsymbol{x}_i - \textbf{e}_i, \: i = 1,2,...,\mathcal{N}, \nonumber
\end{align}
where, $\textbf{e}_i$ is the reconstruction error for a training sample $\boldsymbol{x}_i$. $C$ is the regularization parameter, and $\lambda$ is the graph regularization parameter. $|| \: . \: ||_F$ refers to frobenius norm. After substituting equation (\ref{model_CVdecomp}) in equation (\ref{model_min}), the Langrangian relaxation is found as,
\begin{equation} \label{model_lang}
\mathcal{L} = \frac{1}{2} \left| \left| \left(\boldsymbol{\beta}\right)^T \left(\mathcal{H} \boldsymbol{\mathcal{M}} \left( \mathcal{H} \right)^T+ \lambda \mathbb{I} \right) \: \boldsymbol{\beta} \right| \right|_F^2 + \frac{C}{2} \: \sum_{i=1}^{\mathcal{N}} \left \Vert \textbf{e}_i \right \Vert_2^2 - \sum_{i=1}^{\mathcal{N}} \alpha_i \left( \left( \boldsymbol{\beta} \right)^T \boldsymbol{\mathnormal{h}} \left(\boldsymbol{x}_i \right) - \boldsymbol{x}_i + \textbf{e}_i \right),
\end{equation}
where, $\alpha=\{\alpha_i\}, \: i=1,2,...,\mathcal{N}$, is a Langrangian multiplier. We compute the derivatives as follows:
\begin{align}
& \frac{\partial \mathcal{L}}{\partial \boldsymbol{\beta}} = 0 \implies \boldsymbol{\beta} = \alpha \mathcal{H} \left(\mathcal{H} \boldsymbol{\mathcal{M}} \left( \mathcal{H} \right)^T+ \lambda \mathbb{I} \right)^{-1} \label{model_partbeta}, \\
& \frac{\partial \mathcal{L}}{\partial \textbf{e}_i} = 0 \implies \textbf{E} = \frac{\alpha}{C} \label{model_parterror}, \\
& \frac{\partial \mathcal{L}}{\partial \alpha} = 0 \implies \alpha = C \left( \textbf{X} - \left(\boldsymbol{\beta}\right)^T \mathcal{H} \right) \label{model_partalpha}.
\end{align}
The equation (\ref{model_partalpha}) is substituted in equation (\ref{model_partbeta}) to get,
\begin{equation} \label{model_betaH}
\boldsymbol{\beta} = \mathcal{H} \left( \mathcal{H}\left( \mathcal{H} \right)^T + \frac{\mathcal{H} \boldsymbol{\mathcal{M}} \left( \mathcal{H} \right)^T}{C} + \frac{\lambda}{C} \mathbb{I} \right)^{-1} \textbf{X}. 
\end{equation}
Kernel matrix $\boldsymbol{\mathit{\Omega}}$ is defined as $\boldsymbol{\mathit{\Omega}}=\mathcal{H}\mathcal{H}^T$,  $\mathit{\Omega}_{j,k}=\boldsymbol{\mathnormal{h}}(\boldsymbol{x}_j)\boldsymbol{\mathnormal{h}}(\boldsymbol{x}_k)=K(\boldsymbol{x}_j,\boldsymbol{x}_k), \: j,k = 1,...,\mathcal{N}$. $K$ is a kernel function. Equation (\ref{model_betaH}) is rewritten as,
\begin{equation} \label{model_betaK}
\boldsymbol{\beta} = \left( \boldsymbol{\mathit{\Omega}} + \frac{ \boldsymbol{\mathcal{M}} \boldsymbol{\mathit{\Omega}}}{C} + \frac{\lambda}{C} \mathbb{I} \right)^{-1} \textbf{X}. 
\end{equation}
We express the network output as, 
\begin{equation} \label{model_outK}
\boldsymbol{\widehat{O}} = \begin{bmatrix}
K(\boldsymbol{x},x_1) \\
\vdots\\
K(\boldsymbol{x},x_\mathcal{N})
\end{bmatrix}^T \: \left( \boldsymbol{\mathit{\Omega}} + \frac{ \boldsymbol{\mathcal{M}} \boldsymbol{\mathit{\Omega}}}{C} + \frac{\lambda}{C} \mathbb{I} \right)^{-1} \textbf{X}. 
\end{equation}
Finally, we calculate the threshold ($\theta$) as follows,
\begin{enumerate}
	\item The loss $\boldsymbol{s}$ is calculated using following loss function,
	\begin{equation}
	s_i =\sum_{j=1}^{d}(\widehat{O}_{ij} - x_{ij})^2, \text{  }i=1,2,...,\mathcal{N}.
	\end{equation} 
	\item $\boldsymbol{s}$ is then sorted in decreasing order and is denoted as, $\boldsymbol{s}_{dec}$. The threshold ($\theta$) is then calculated as,
	\begin{equation} \label{model_thresh}
	\theta=\boldsymbol{s}_{dec}(\floor{\delta * \mathcal{N}}),    
	\end{equation}
	where, $\delta$ is the fraction of dismissal.
\end{enumerate}
For a test sample $\boldsymbol{x}_t$, we determine the network output $\widehat{O}_t$,
\begin{equation} \label{model_outK_test}
\widehat{O}_t = \begin{bmatrix}
K(\boldsymbol{x}_t,\boldsymbol{x}_1) \\
\vdots\\
K(\boldsymbol{x}_t,\boldsymbol{x}_\mathcal{N})
\end{bmatrix}^T  \boldsymbol{\beta}.
\end{equation}
The loss ($s_t$) is then calculated as, $s_t=\sum_{j=1}^{d}\left(\widehat{O}_{tj} - x_{tj}\right)^2$. Finally, classification is done using following decision rule,
\begin{align} \label{model_decision}
sign(\theta-s_t) &= 1, \: \: \: \boldsymbol{x}_t \: belongs \: to \: target \: class, \\
& \: -1, \: \: \: \boldsymbol{x}_t \: belongs \: to \: outlier \: class. \nonumber
\end{align}

\noindent
Algorithm \ref{algo_VAAKELM} briefly provides the implementation steps for the proposed classifier.

\begin{algorithm}
	\textbf{Given:} \newline
	Training dataset: \textbf{X}, Regularization parameter: C, Graph regularization parameter: $\lambda$, Fraction of dismissal: $\delta$
	\newline
	\textbf{Training:}
	\begin{algorithmic}[1]
		\item Calculate kernel matrix $\boldsymbol{\mathit{\Omega}}$.
		\item Calculate output weight $\boldsymbol{\beta}$ using (\ref{model_betaK}).
		\item Calculate network output $\boldsymbol{\widehat{O}}$ using (\ref{model_outK}). 
		\item Calculate threshold $\theta$ using (\ref{model_thresh}).
	\end{algorithmic} 
	\textbf{Testing:}
	\begin{algorithmic}[1]
		\item For test sample $\boldsymbol{x}_t$, calculate network output $\widehat{O}_t$ using (\ref{model_outK_test}).
		\item Classify $\boldsymbol{x}_t$ using (\ref{model_decision}).
	\end{algorithmic}
	\caption{\textit{VAAKELM}}\label{algo_VAAKELM}
\end{algorithm}

\section{Experiments} \label{sec_exp}
We conduct experiments on 15 small-size and 10 medium-size UCI benchmark datasets. The datasets have been downloaded from the UCI Machine Learning Repository \cite{Dua:2019}, and the website of TU Delft\footnote{\url{http://homepage.tudelft.nl/n9d04/occ/}}, made available by Tax and Duin \cite{OCCDataset} in the preprocessed form for OCC. Tax and Duin \cite{OCCDataset} obtained the one-class datasets from the multi-class datasets by taking one of the classes as target and the rest of the classes as outliers. We have followed the same approach. We normalize all the features with a mean 0 and standard deviation 1 using z-score. 80\% of the target and outlier class samples are used for 5-fold cross-validation, and the remaining 20\% is used as the test set. It is important to note that we use samples from only the target class to train the model. The optimal parameters are selected using 5-fold cross-validation from a range of values. The regularization parameter $C$ is selected from the range \{$2^{-5}, 2^{-4},...., 2^{5}$\}. The graph regularization parameter $\lambda$ is taken as 1 in all the experiments. The number of clusters $k$ for k-means clustering is selected from the range \{$1, ...., 10$\}. The fraction of dismissal of outliers $\delta$ is selected from the range \{$1\%, 5\%, 10\%$\}. All the classifiers employ the Radial Basis Function (RBF) kernel, which can be calculated for data points $\bm{x_i}$ and $\bm{x_j}$ as follows:
\begin{equation}\label{Eq:RBF_Kernel}
\begin{aligned}
\bm{k(x_i, x_j)} = exp\left ( -\frac{\left \| \bm{x_i}-\bm{x_j} \right \|_2^2}{2\sigma^2} \right )
\end{aligned},
\end{equation}
where, we use the mean of the euclidean distance across different training samples to obtain $\sigma$. We keep the same experimental setup across all the classifiers to ensure a fair comparison.

The performance of \textit{VAAKELM} is compared with 13 existing one-class classifiers, namely, One Class Random Forests (OCRF) \cite{desir2013one}, Principal Component Analysis (PCA) \cite{bishop1995neural}, Naive Parzen density estimation \cite{duin1976choice}, k-means \cite{jiang2001two}, k-Nearest Neighbor (k-NN) \cite{knorr2000distance}, Autoencoder neural network or Multi-layer Perceptron (MLP) \cite{tax2002one}, k-centers \cite{hochbaum1985best}, Support Vector Data Description (SVDD) \cite{tax2004support}, One Class Support Vector Machine (OCSVM) \cite{scholkopf2001estimating}, Minimum Spanning Tree based one-class classifier (MST) \cite{juszczak2009minimum}, OCKELM \cite{leng2015one}, VOCKELM \cite{mygdalis2016one}, and AAKELM \cite{gautam2017construction}. The motivation behind choosing the existing one-class classifiers for comparison purpose is based upon the fact that they have been used as benchmark classifiers frequently in the past \cite{desir2013one,juszczak2009minimum} and are regarded as the standard classifiers in the field of OCC \cite{pimentel2014review}. The implementations of the classifiers are taken from ddtools \cite{Ddtools2018}. OCSVM is implemented using the LIBSVM library \cite{CC01a}.

\subsection{Performance metrics}
\noindent	We have adopted the following metrics for performance evaluation,
\begin{equation} \label{accu}
Accuracy = \frac{TP+TN}{TP+TN+FP+FN},
\end{equation}
\begin{equation} \label{prec}
Precision \: (P) = \frac{TP}{TP+FP},
\end{equation}
\begin{equation} \label{rec}
Recall \: (R) = \frac{TP}{TP+FN},
\end{equation}
\begin{equation} \label{f1score}
F_1 \: score = \frac{2 \: P.R}{P+R},
\end{equation}
\begin{equation} \label{gmean}
G-mean = \sqrt{P. R}.
\end{equation}

Above, \textit{FN}, \textit{FP}, \textit{TN}, and \textit{TP} represent false negative, false positive, true negative, and true positive, respectively. Accuracy denotes the fraction of all correct measurements. Precision reflects the fraction of correct positive measurements among all the predicted positives. Recall indicates the fraction of correct positive measurements among the actual positives. F$_1$ score and G-mean are the harmonic mean and geometric mean of precision and recall, respectively.

In the case of imbalanced datasets, it is possible to obtain good accuracy by classifying any given sample to the majority class. Taking an example, suppose we have an imbalanced dataset where 90 samples belong to the positive class, and 10 samples belong to the negative class. Now, a model may classify all the negative class samples incorrectly to the positive class. In such a case, the accuracy will be determined as 90\%, even if all the negative class samples are incorrectly classified. Hence, accuracy fails to give an unbiased score for the performance of a model when the data is imbalanced. Precision and recall provide a better understanding of the efficiency of a model in such cases. In order to obtain an equilibrium between precision and recall, researchers
mostly use F$_1$ score and g-mean \cite{manevitz2007one,leng2015one,mygdalis2016one,iosifidis2017one} when the data is imbalanced. In this paper, we use the F$_1$ score as the first evaluation metric as most of the datasets that we have used for the experiments are imbalanced in nature. Since we have to compare multiple classifiers on various datasets, we compute the mean of all F$_1$ scores ($\eta_{\textbf{F}_1}$) over all the datasets by taking inspiration from an existing work \cite{fernandez2014we}. We consider $\eta_{\textbf{F}_1}$ as the final evaluation measure to rank the classifiers as per their performance. For reference, we also present the results based on accuracy, g-mean, precision, and recall metrics.

\subsection{Experimental Results}
The performance of \textit{VAAKELM} is compared with 13 existing one-class classifiers on the basis of $\eta_{\textbf{F}_1}$. Further, we divide this section into two parts. In Section \ref{sec_exp_small}, we discuss experimental results on small-size datasets. In Section \ref{sec_exp_medium}, we discuss experimental results on medium-size datasets. 

\subsubsection{Experiments on small-size datasets}\label{sec_exp_small}

\begin{table}[b]
	\centering
	\fontsize{6.5}{8}\selectfont
	\begin{tabular}{l l r r r r l} 
		\hline
		S.no. & Datasets             & \#Total Samples & \#Target & \#Outlier & \#Features & Target Class \\
		\hline
		1     & Biomed               & 194             & 127      & 67        & 5          & Healthy             \\
		2     & Breast Cancer        & 699             & 241      & 458       & 9          & Malignant           \\
		3     & Cardiotocography     & 2126            & 176      & 1950      & 22         & Pathologic          \\
		4     & Colposcopy           & 97              & 82       & 15        & 62         & Good                \\
		5     & Cryotherapy          & 90              & 48       & 42        & 6          & 1                   \\
		6     & Diabetic Retinopathy & 1151            & 540      & 611       & 19         & Normal              \\
		7     & Ecoli                & 336             & 52       & 284       & 7          & Periplasm           \\
		8     & Heart Cleveland      & 297             & 160      & 137       & 13         & Absent              \\
		9     & Heart Statlog        & 270             & 120      & 150       & 13         & Present             \\
		10    & Imports              & 159             & 71       & 88        & 25         & Low Risk            \\
		11    & Sonar                & 208             & 97       & 111       & 60         & Rocks               \\
		12    & Survival             & 306             & 225      & 81        & 3          & Greater than 5 year \\
		13    & Vowel                & 528             & 48       & 480       & 10         & 0                   \\
		14    & Waveform             & 900             & 300      & 600       & 21         & 1                   \\
		15    & Wine                 & 178             & 71       & 107       & 13         & 2           \\
		\hline 
	\end{tabular}
	\caption{Small-size one-class datasets specifications.}
	\label{tab:spec_small}
\end{table}

\begin{figure*}
	\begin{center}
		\begin{subfigure}{0.8\textwidth}
			\includegraphics[width=\linewidth]{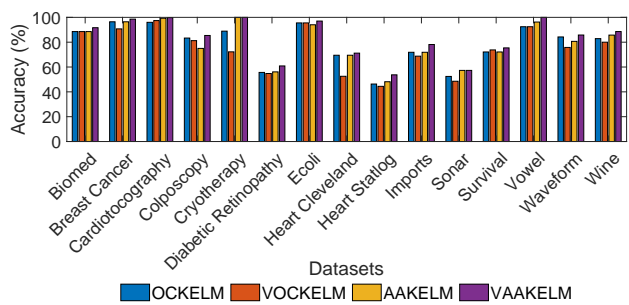}
			\caption{Accuracy}
		\end{subfigure}
		
		\begin{subfigure}{0.8\textwidth}
			\includegraphics[width=\linewidth]{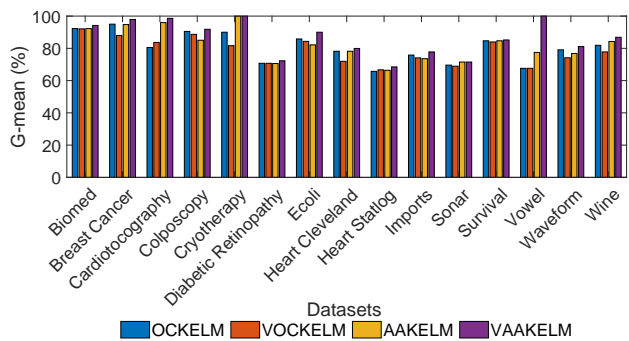}
			\caption{G-mean}
		\end{subfigure}
		
		\begin{subfigure}{0.8\textwidth}
			\includegraphics[width=\linewidth]{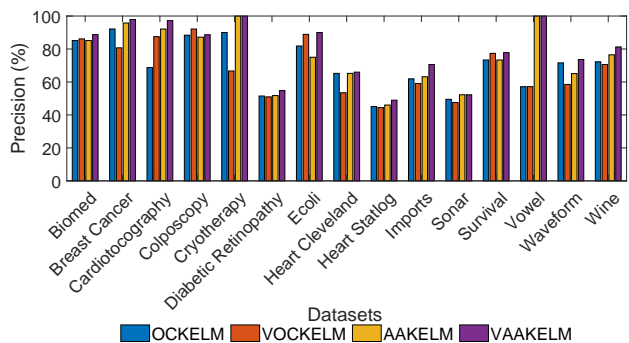}
			\caption{Precision}
		\end{subfigure}
		
		\begin{subfigure}{0.8\textwidth}
			\includegraphics[width=\linewidth]{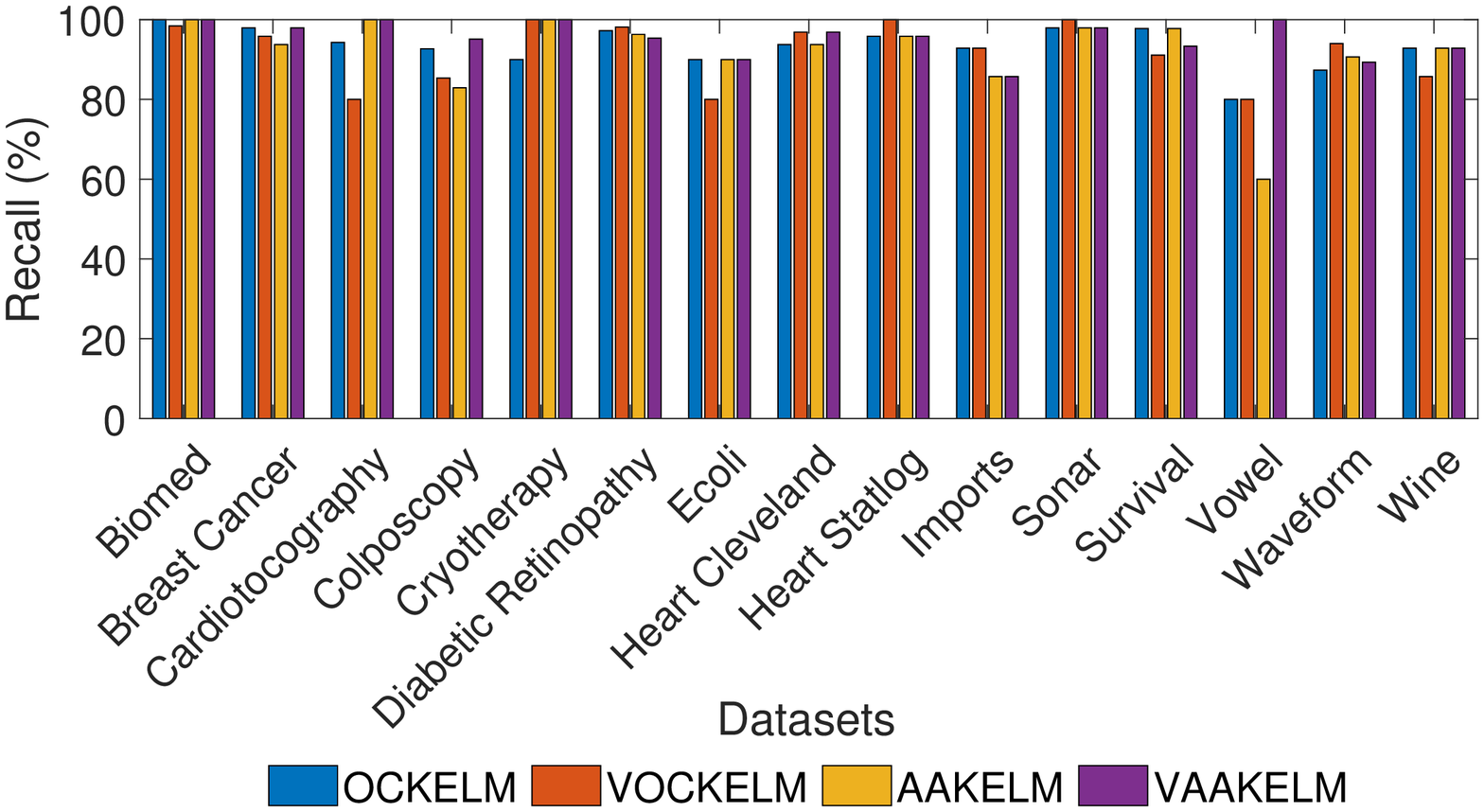}
			\caption{Recall}
		\end{subfigure}
		\caption{Examination of Accuracy, G-mean, Precision and Recall for small-size one-class datasets.}
		\label{fig:plotmetrics_small}
	\end{center}
\end{figure*}

We conduct experiments on 15 small-size one-class datasets. We provide the specifications of these datasets in Table \ref{tab:spec_small}. We compare the performance of \textit{VAAKELM} with various existing one-class classifiers in Table \ref{tab:F1} based on F$_1$ score. The first row in the table lists the name of classifiers, the first column lists the datasets, and the last row lists the $\eta_{\textbf{F}_1}$ value for
each one-class classifier. We consider $\eta_{\textbf{F}_1}$ as the final evaluation measure to rank the classifiers as per their performance. \textit{VAAKELM} obtains the highest $\eta_{\textbf{F}_1}$ (highlighted in bold red) as compared to the existing one-class classifiers, with a significant improvement of 7.47\% in comparison to non-kernel-based classifiers and 4.53\% in comparison to other kernel-based classifiers. Further, it can be noted that \textit{VAAKELM} obtains the highest F$_1$ score for all 15 datasets (highlighted in bold). \textit{VAAKELM} achieves this by reducing the intra-class variance and leveraging kernel autoencoder to learn essential features from the input data. Thus, it is evident that \textit{VAAKELM} performs better than the existing one-class classifiers for small-size datasets, and can be used as a viable alternative for OCC.
For reference, we also present the experimental results based on accuracy,
g-mean, precision, and recall metrics in Figure \ref{fig:plotmetrics_small}. \textit{VAAKELM} scores the highest accuracy, g-mean, and precision for an overwhelming 15, 15, and 14 datasets, respectively. 

\begin{table*}[t]
	\centering
	\fontsize{5}{8}\selectfont
	\begin{tabular}{lllllllllllllll}
		\hline
		\multirow{2}{*}{}      & \multirow{2}{*}{\makecell{OCRF\\\cite{desir2013one}}} & \multirow{2}{*}{\makecell{Naive\\Parzen\\\cite{duin1976choice}}} & \multirow{2}{*}{\makecell{k-means\\\cite{jiang2001two}}} & \multirow{2}{*}{\makecell{k-NN\\\cite{knorr2000distance}}} & \multirow{2}{*}{\makecell{Auto\\encoder\\\cite{tax2002one}}} & \multirow{2}{*}{\makecell{PCA\\\cite{bishop1995neural}}} & \multirow{2}{*}{\makecell{MST\\\cite{juszczak2009minimum}}} & \multicolumn{1}{l|}{\multirow{2}{*}{\makecell{kCentre\\\cite{hochbaum1985best}}}} & \multicolumn{6}{c}{Kernel-based one-class classifiers}                   \\
		\cline{10-15}
		&                       &       &           &        &          &      &            &          \multicolumn{1}{l|}{}            & \makecell{OCSVM\\\cite{scholkopf2001estimating}} & \makecell{SVDD\\\cite{tax2004support}}  & \makecell{OCKELM\\\cite{leng2015one}} & \makecell{VOCKLEM\\\cite{mygdalis2016one}} & \makecell{AAKELM\\\cite{gautam2017construction}} & \makecell{VAAKLEM}        \\
		\hline
		Biomed                 & 79.25                 & 93.75                                                                                        & 91.18                    & 89.86                & 88.37                        & 94.03                & 91.18                & 88.57                    & 91.85 & 90.63 & 91.97  & 91.85   & 91.97  & \textbf{94.03} \\
		Breast Cancer          & 51.06                 & 87.23                                                                                        & 95.92                    & 51.06                & 95.83                        & 51.06                & 69.92                & 83.81                    & 95.83 & 94.74 & 94.95  & 87.62   & 94.74  & \textbf{97.92} \\
		Cardiotocography       & 15.22                 & 41.18                                                                                        & 34.68                    & 42.31                & 92.31                        & 36.57                & 47.41                & 30.77                    & 19.22 & 20.29 & 79.52  & 83.58   & 95.89  & \textbf{98.59} \\
		Colposcopy             & 87.06                 & 81.08                                                                                        & 89.66                    & 90.91                & 88.37                        & 86.75                & 90.91                & 89.66                    & 86.75 & 82.05 & 90.48  & 88.61   & 85     & \textbf{91.76} \\
		Cryotherapy            & 71.43                 & 95.24                                                                                        & 73.68                    & 83.33                & 84.21                        & 90                   & 86.96                & 85.71                    & 90    & 77.78 & 90     & 80      & 100    & \textbf{100}   \\
		Diabetic   Retinopathy & 63.91                 & 69.6                                                                                         & 66.03                    & 66.45                & 68.67                        & 66.46                & 67.1                 & 65.46                    & 65.64 & 66.46 & 67.31  & 67.09   & 67.31  & \textbf{69.6}  \\
		Ecoli                  & 25.97                 & 90                                                                                           & 85.71                    & 78.26                & 75                           & 34.15                & 85.71                & 43.48                    & 58.33 & 60    & 85.71  & 84.21   & 81.82  & \textbf{90}    \\
		Heart   Cleveland      & 70.33                 & 74.67                                                                                        & 69.05                    & 66.67                & 71.91                        & 68.89                & 69.66                & 68.24                    & 67.44 & 65.06 & 76.92  & 68.89   & 76.92  & \textbf{78.48} \\
		Heart Statlog          & 61.54                 & 61.29                                                                                        & 60.53                    & 61.54                & 62.34                        & 60.53                & 60.27                & 63.16                    & 60.27 & 59.16 & 61.33  & 61.54   & 62.16  & \textbf{64.79} \\
		Imports                & 65.12                 & 68.75                                                                                        & 63.64                    & 56.41                & 59.46                        & 68.42                & 66.67                & 55.81                    & 70.59 & 68.75 & 74.29  & 72.22   & 72.73  & \textbf{77.42} \\
		Sonar                  & 66.67                 & 63.72                                                                                        & 64                       & 63.95                & 67.63                        & 62.02                & 63.95                & 67.65                    & 64.18 & 63.25 & 65.73  & 64.43   & 68.12  & \textbf{68.12} \\
		Survival               & 84.9                  & 81.25                                                                                        & 84.31                    & 82.35                & 82.69                        & 83.81                & 82.35                & 83.81                    & 83.81 & 82.69 & 83.81  & 83.67   & 83.81  & \textbf{84.9}  \\
		Vowel                  & 17.24                 & 75                                                                                           & 82.35                    & 82.35                & 83.33                        & 47.62                & 94.74                & 82.35                    & 28.57 & 41.67 & 66.67  & 66.67   & 75     & \textbf{100}   \\
		Waveform               & 50                    & 78.26                                                                                        & 73.6                     & 73.25                & 70.8                         & 72.33                & 73.33                & 71.2                     & 79.14 & 75.78 & 78.68  & 72.12   & 75.77  & \textbf{80.72} \\
		Wine                   & 57.14                 & 72                                                                                           & 78.79                    & 82.35                & 80                           & 55.32                & 82.35                & 78.79                    & 80    & 75    & 81.25  & 77.42   & 83.87  & \textbf{86.67} \\
		\hline
		$\eta_{\textbf{F}_1}$             & 57.79                 & 75.53                                                                                        & 74.21                    & 71.40                & 78.06                        & 65.20                & 75.50                & 70.56                    & 69.44 & 68.22 & 79.24  & 76.66   & 81.01  & \textcolor{red}{\textbf{85.53}} \\
		\hline
	\end{tabular}
	\caption{F$_1$ score comparisons for different one-class classifiers for small-size one-class datasets.}
	\label{tab:F1}
\end{table*}

\begin{figure*}
	\begin{center}
		\begin{subfigure}{0.3\textwidth}
			\includegraphics[width=\linewidth]{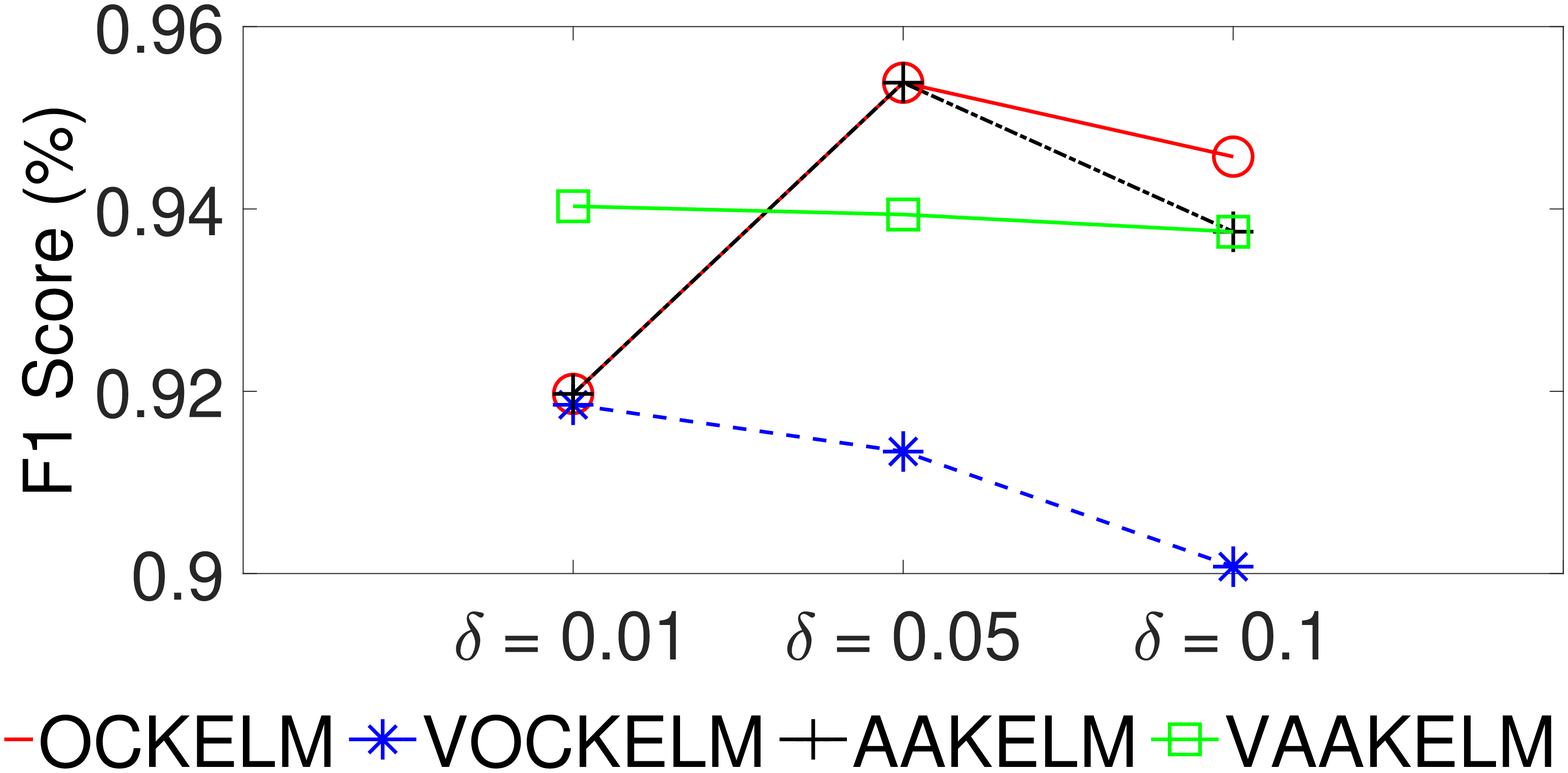}
			\caption{Biomed}
		\end{subfigure}
		\begin{subfigure}{0.3\textwidth}
			\includegraphics[width=\linewidth]{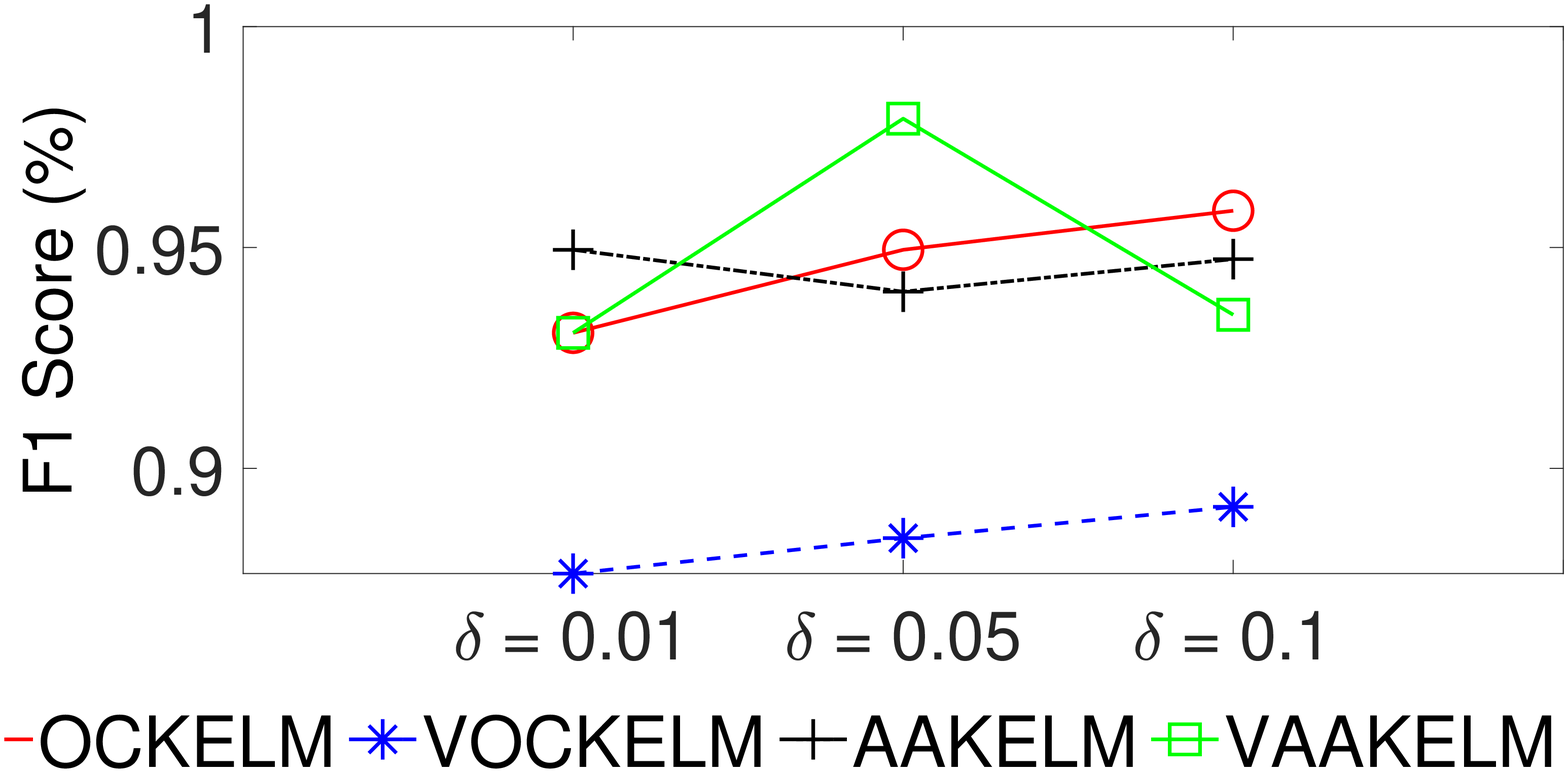}
			\caption{Breast Cancer}
		\end{subfigure}
		\begin{subfigure}{0.3\textwidth}
			\includegraphics[width=\linewidth]{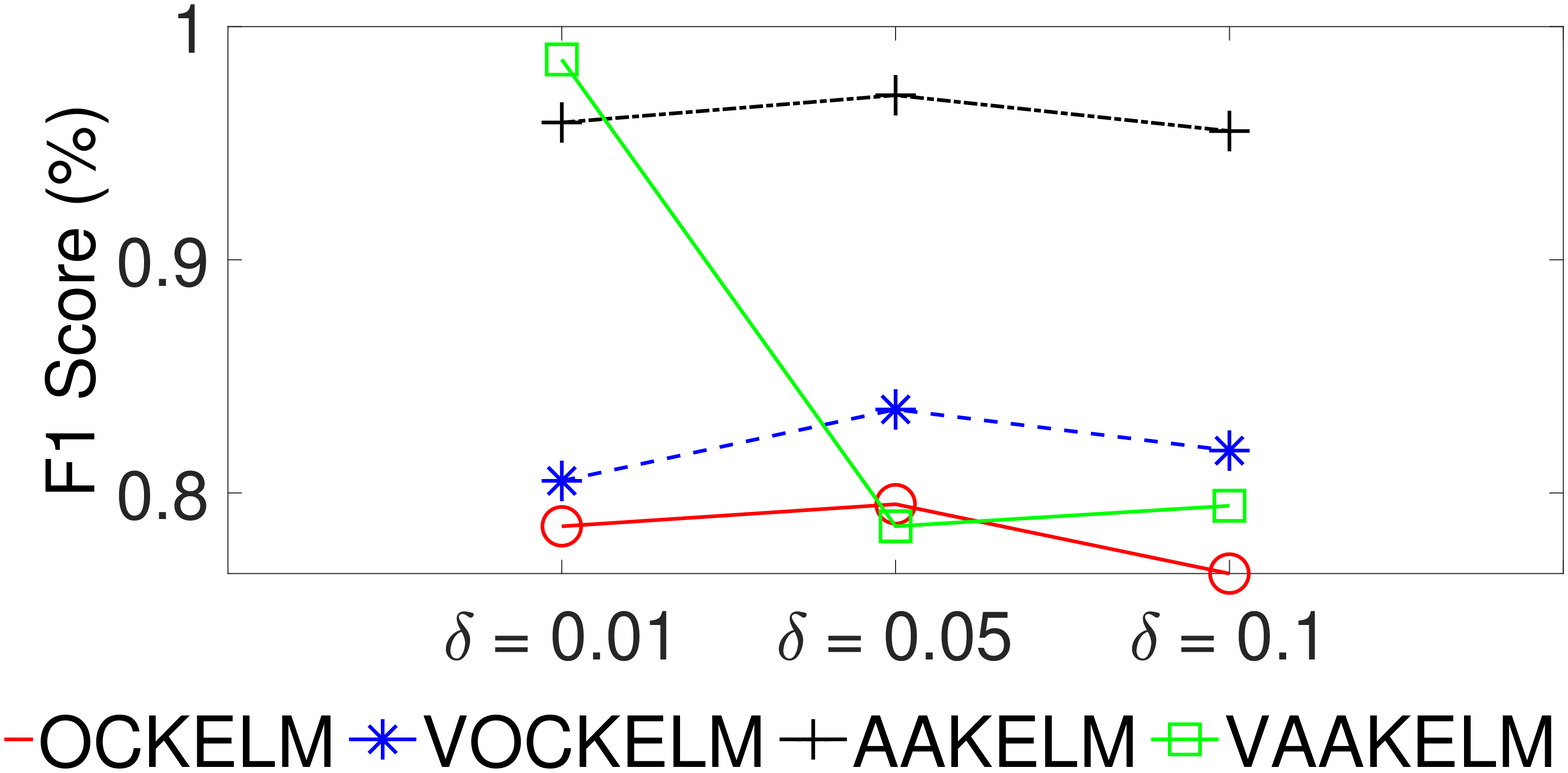}
			\caption{Cardiotocography}
		\end{subfigure}
		
		\begin{subfigure}{0.3\textwidth}
			\includegraphics[width=\linewidth]{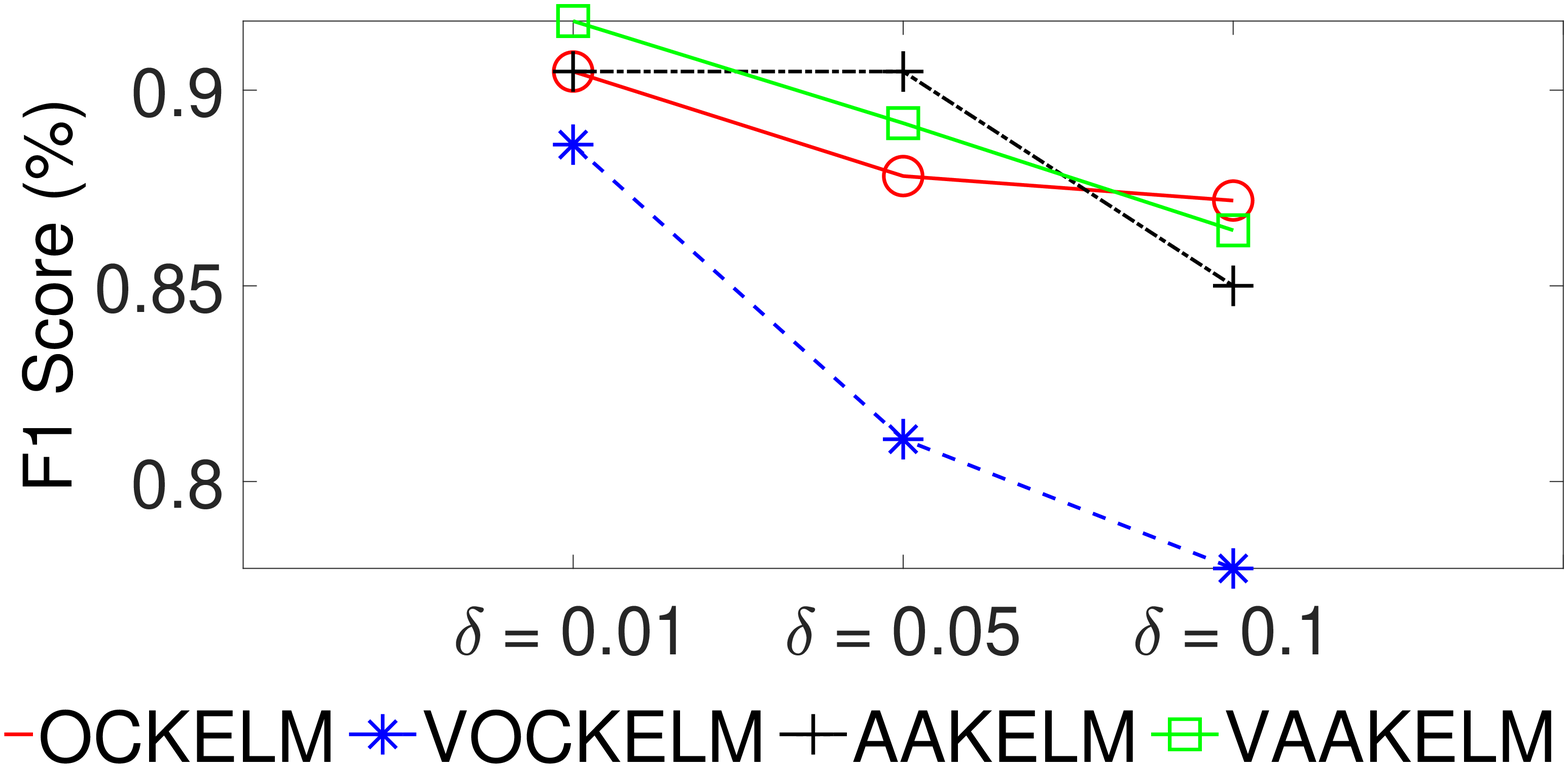}
			\caption{Colposcopy}
		\end{subfigure}
		\begin{subfigure}{0.3\textwidth}
			\includegraphics[width=\linewidth]{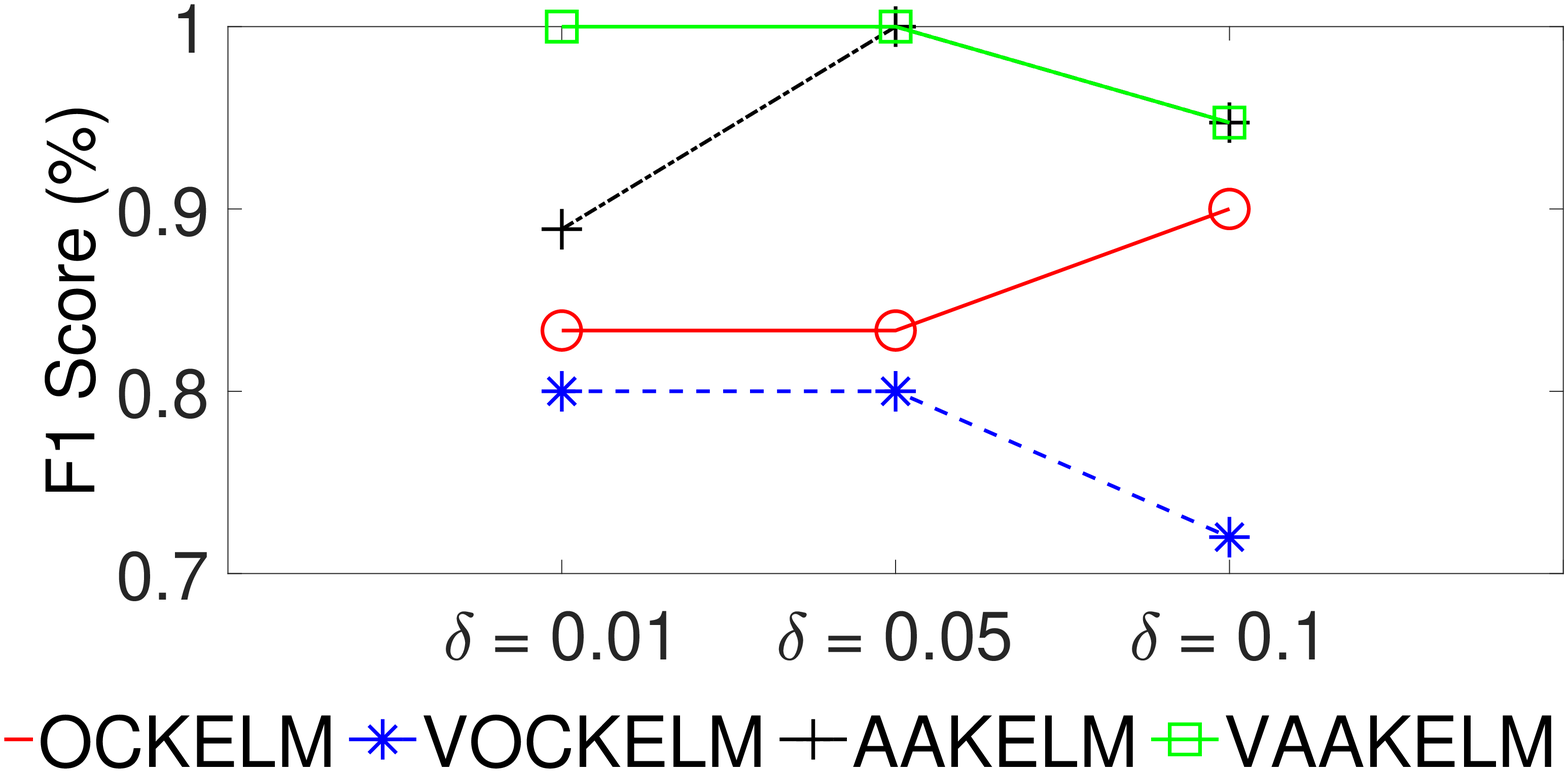}
			\caption{Cryotherapy}
		\end{subfigure}
		\begin{subfigure}{0.3\textwidth}
			\includegraphics[width=\linewidth]{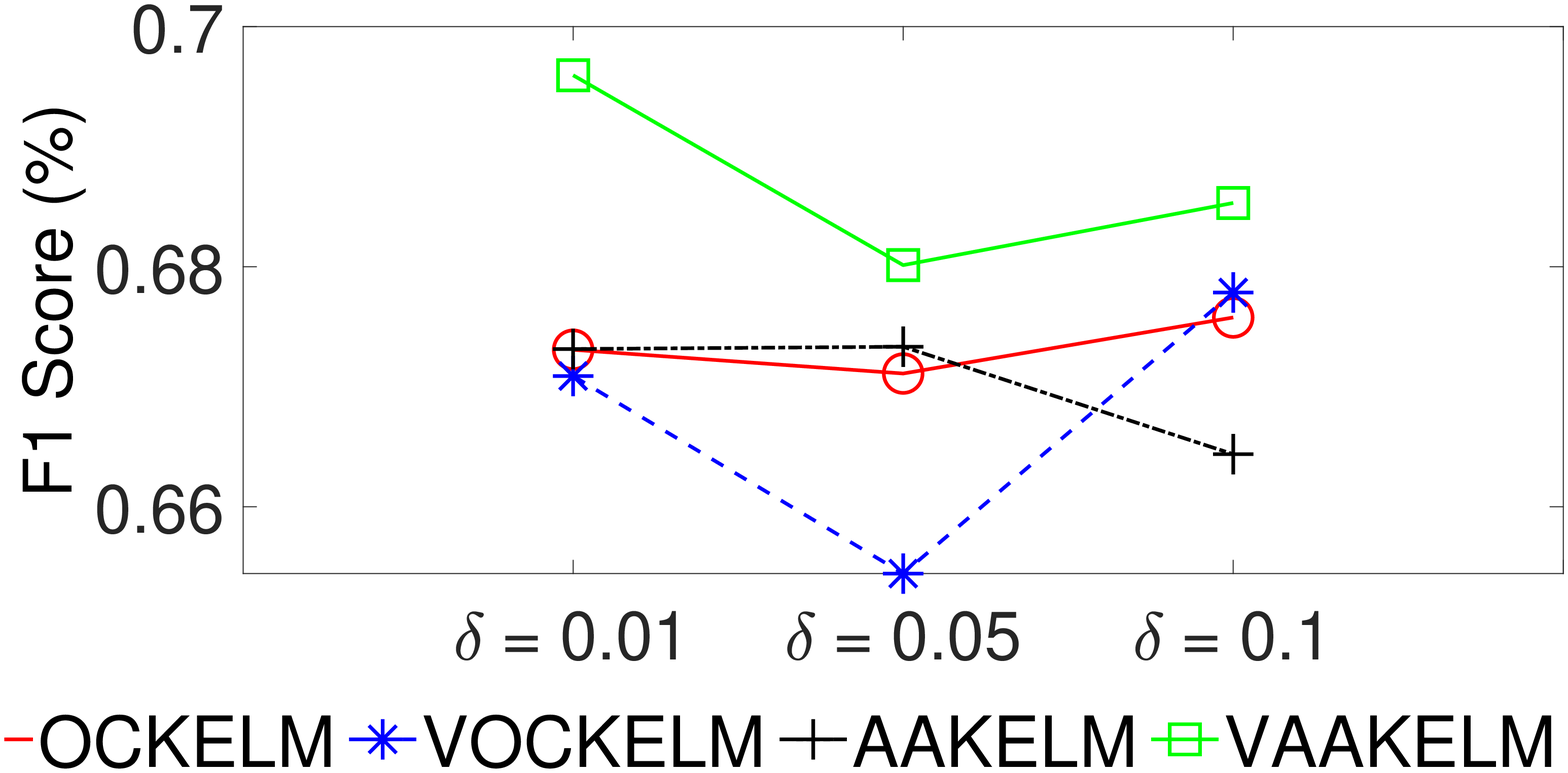}
			\caption{Diabetic Retinopathy}
		\end{subfigure}
		
		\begin{subfigure}{0.3\textwidth}
			\includegraphics[width=\linewidth]{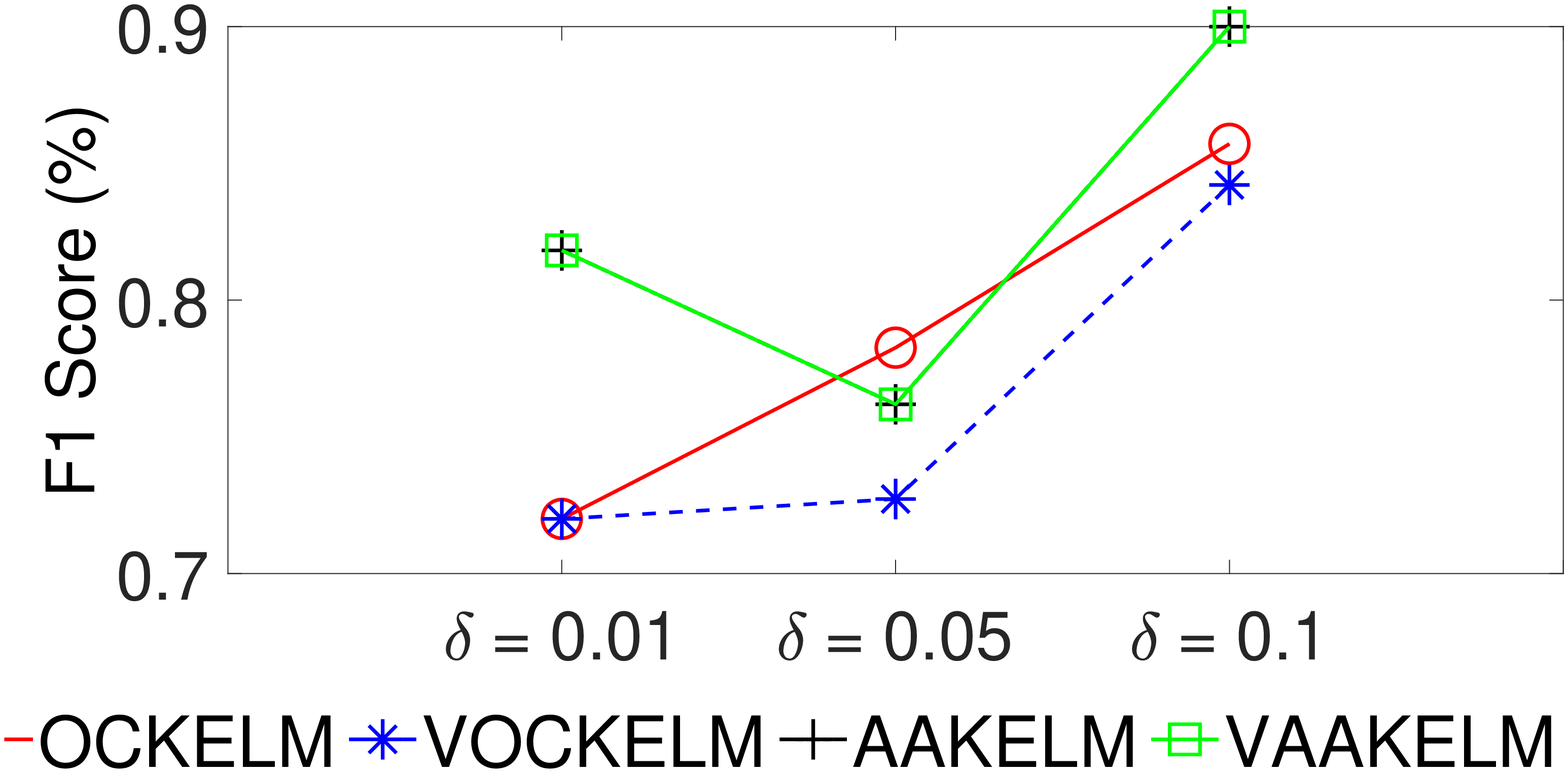}
			\caption{Ecoli}
		\end{subfigure}
		\begin{subfigure}{0.3\textwidth}
			\includegraphics[width=\linewidth]{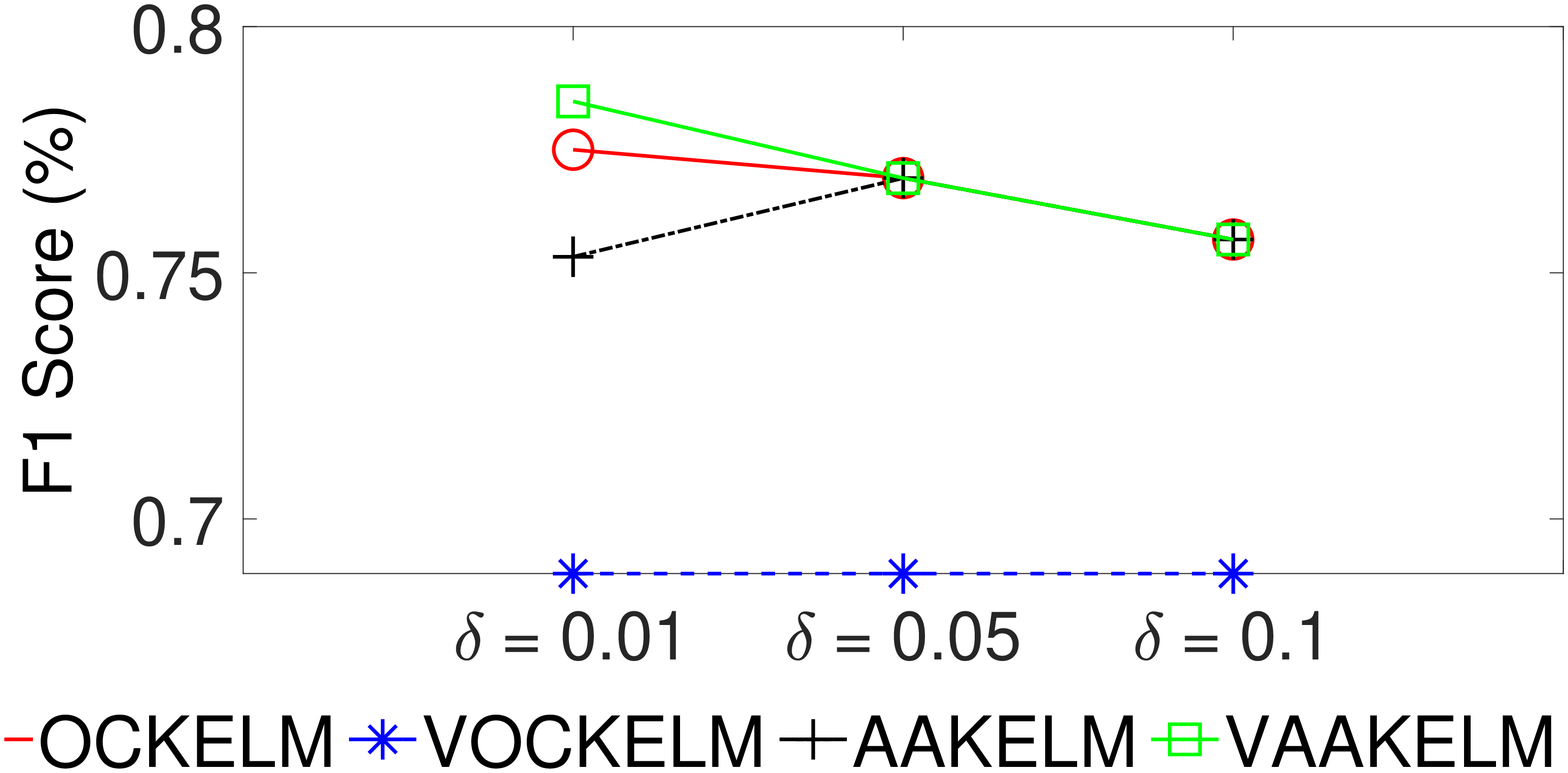}
			\caption{Heart Cleveland}
		\end{subfigure}
		\begin{subfigure}{0.3\textwidth}
			\includegraphics[width=\linewidth]{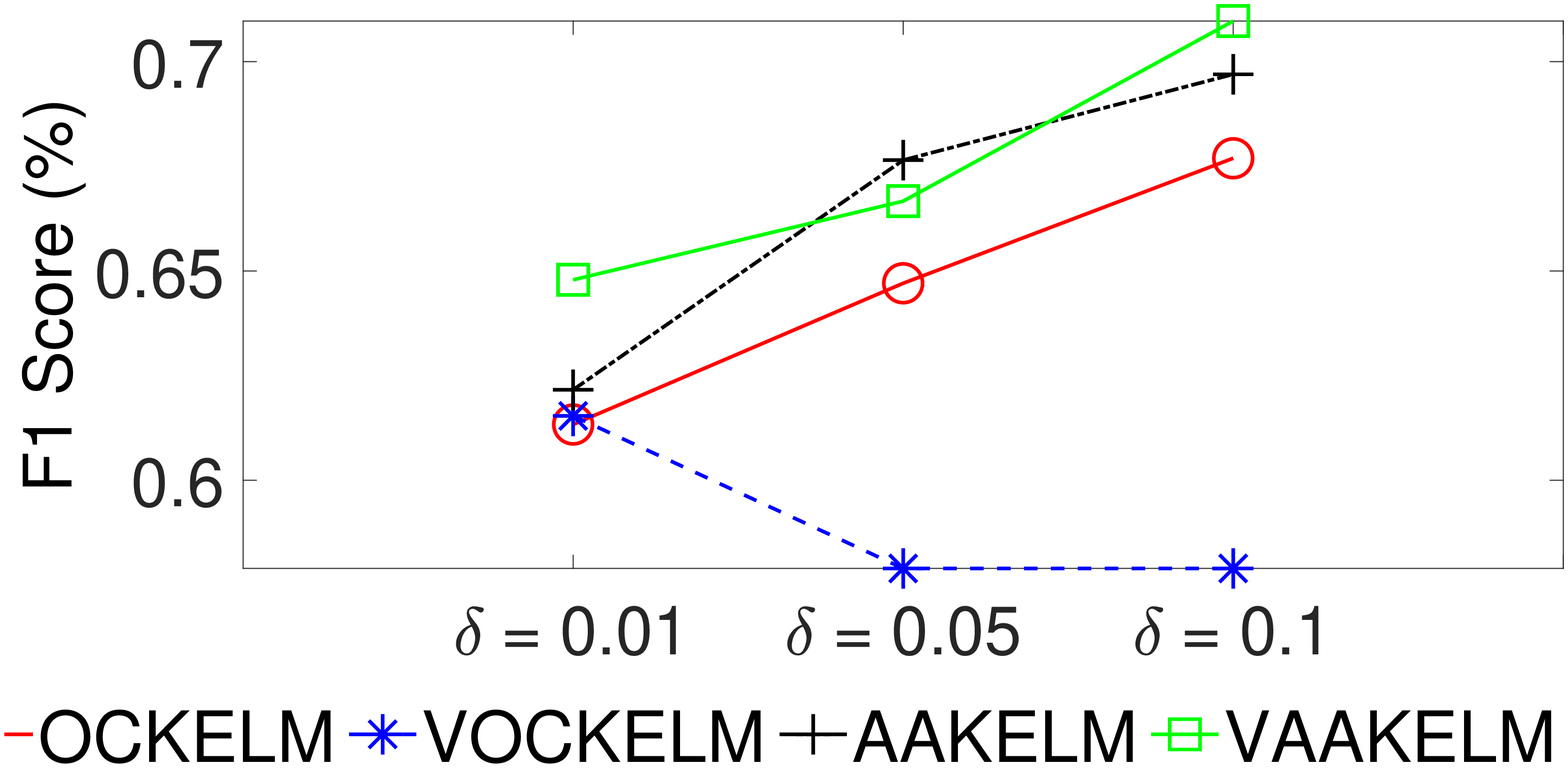}
			\caption{Heart Statlog}
		\end{subfigure}
		
		\begin{subfigure}{0.3\textwidth}
			\includegraphics[width=\linewidth]{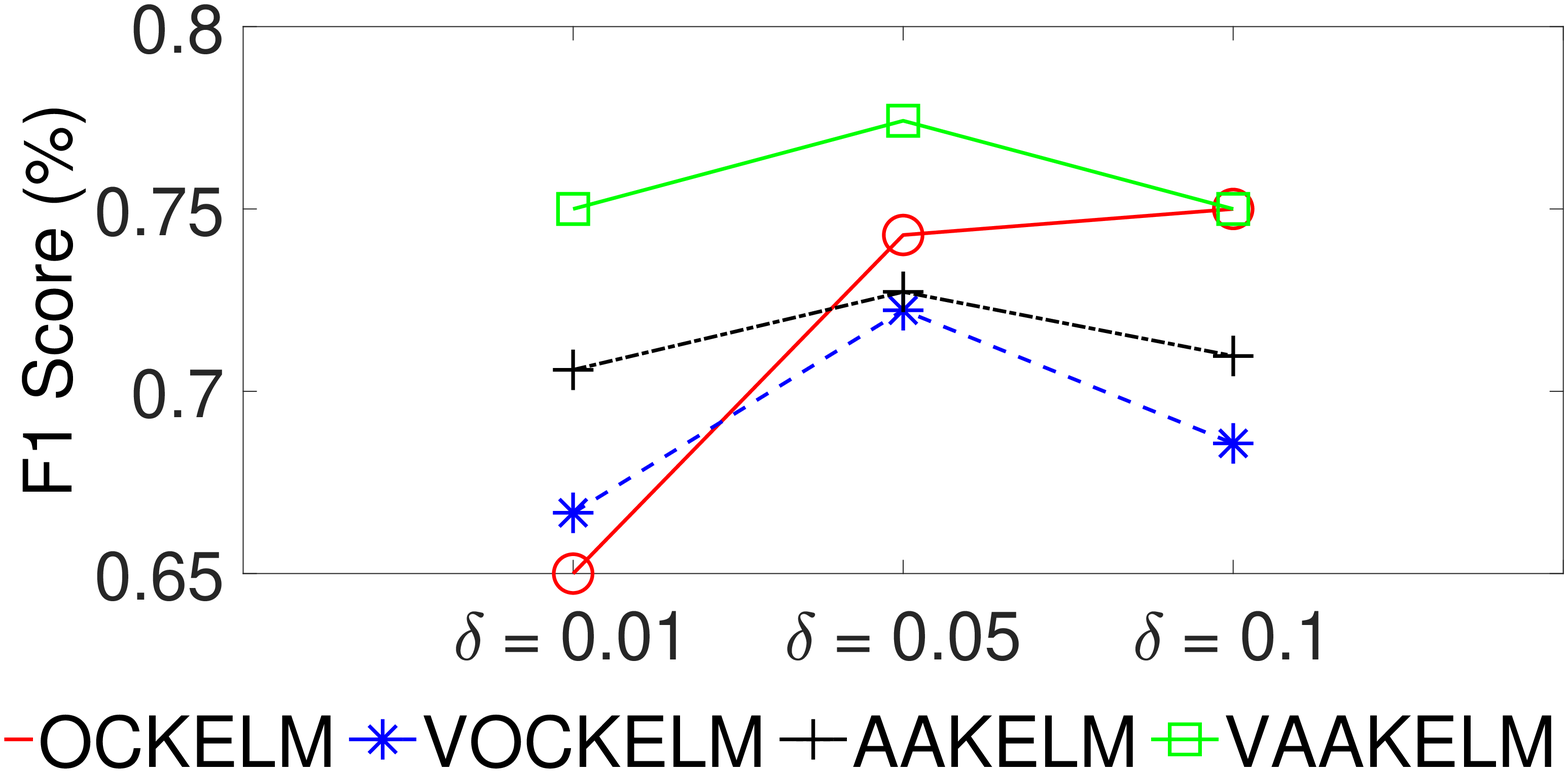}
			\caption{Imports}
		\end{subfigure}
		\begin{subfigure}{0.3\textwidth}
			\includegraphics[width=\linewidth]{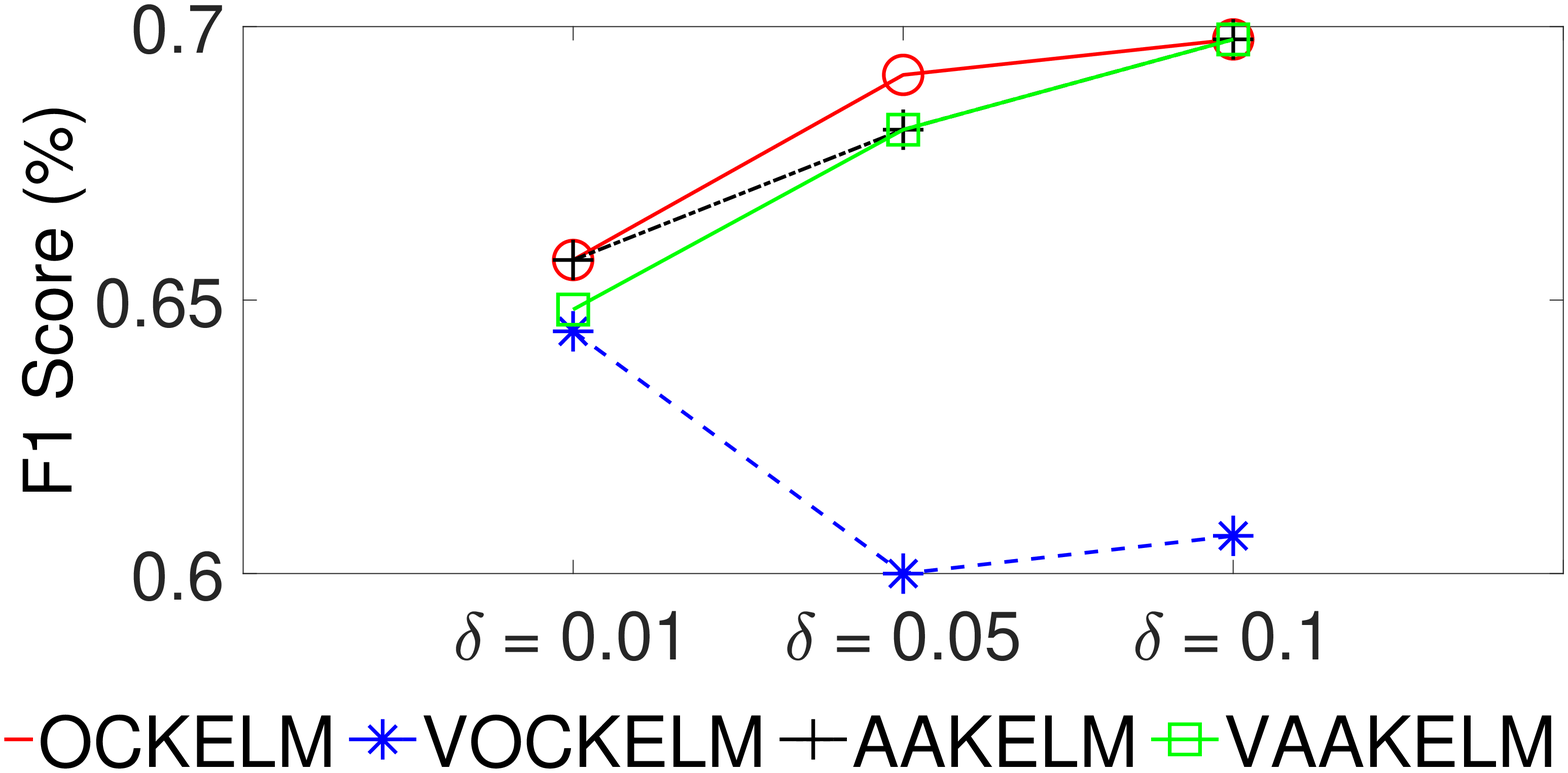}
			\caption{Sonar}
		\end{subfigure}
		\begin{subfigure}{0.3\textwidth}
			\includegraphics[width=\linewidth]{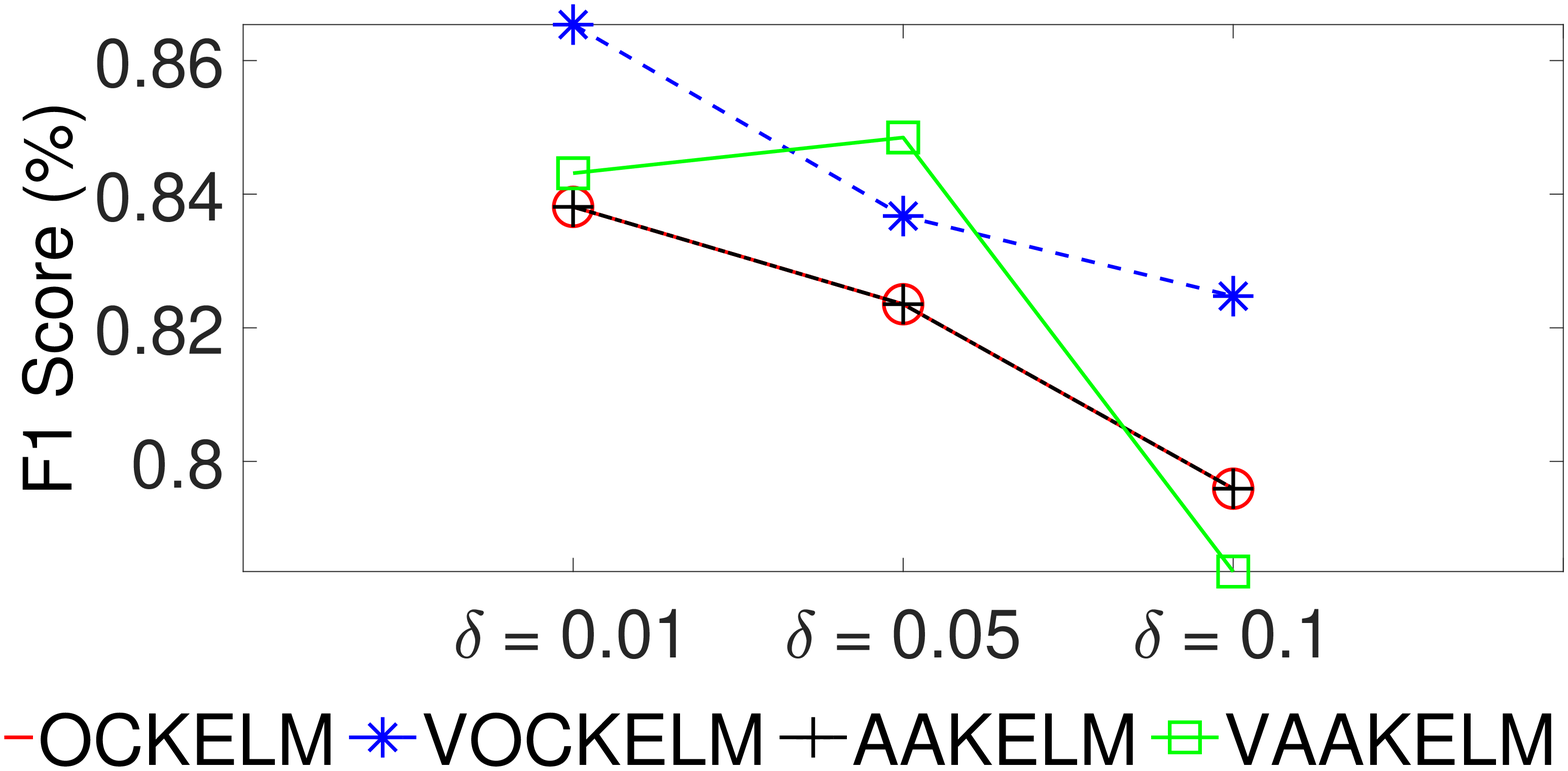}
			\caption{Survival}
		\end{subfigure}
		
		\begin{subfigure}{0.3\textwidth}
			\includegraphics[width=\linewidth]{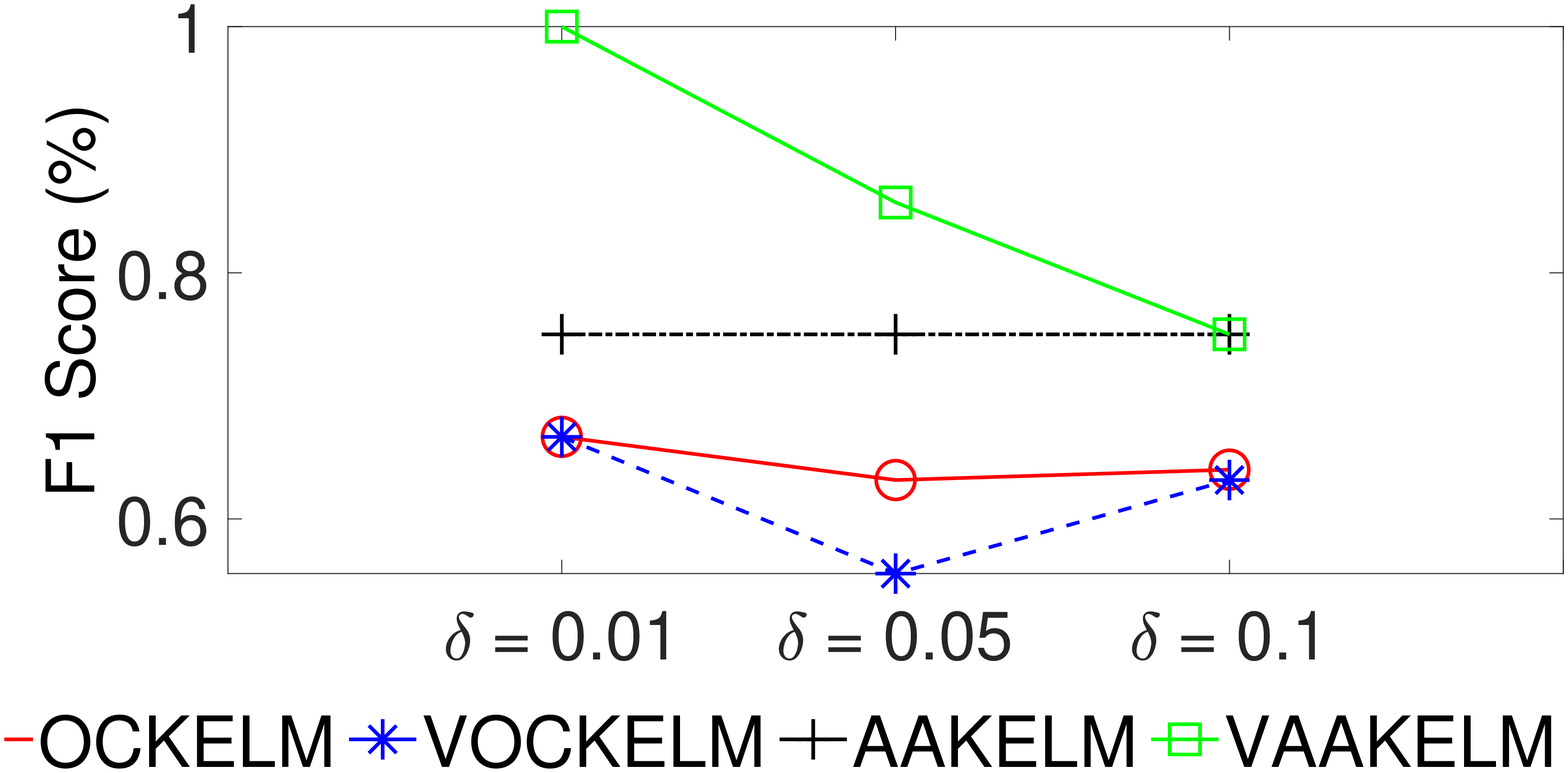}
			\caption{Vowel}
		\end{subfigure}
		\begin{subfigure}{0.3\textwidth}
			\includegraphics[width=\linewidth]{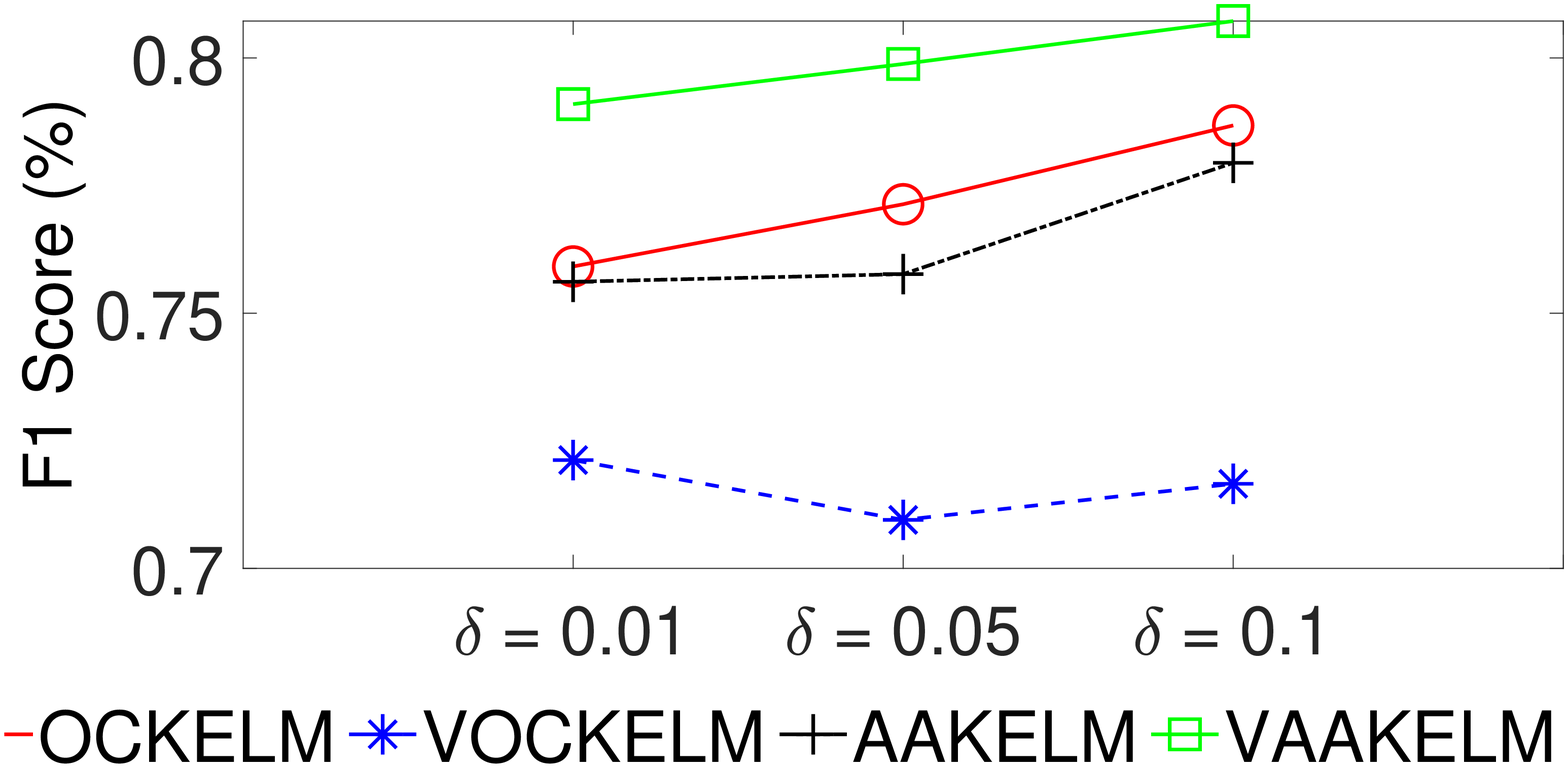}
			\caption{Waveform}
		\end{subfigure}
		\begin{subfigure}{0.3\textwidth}
			\includegraphics[width=\linewidth]{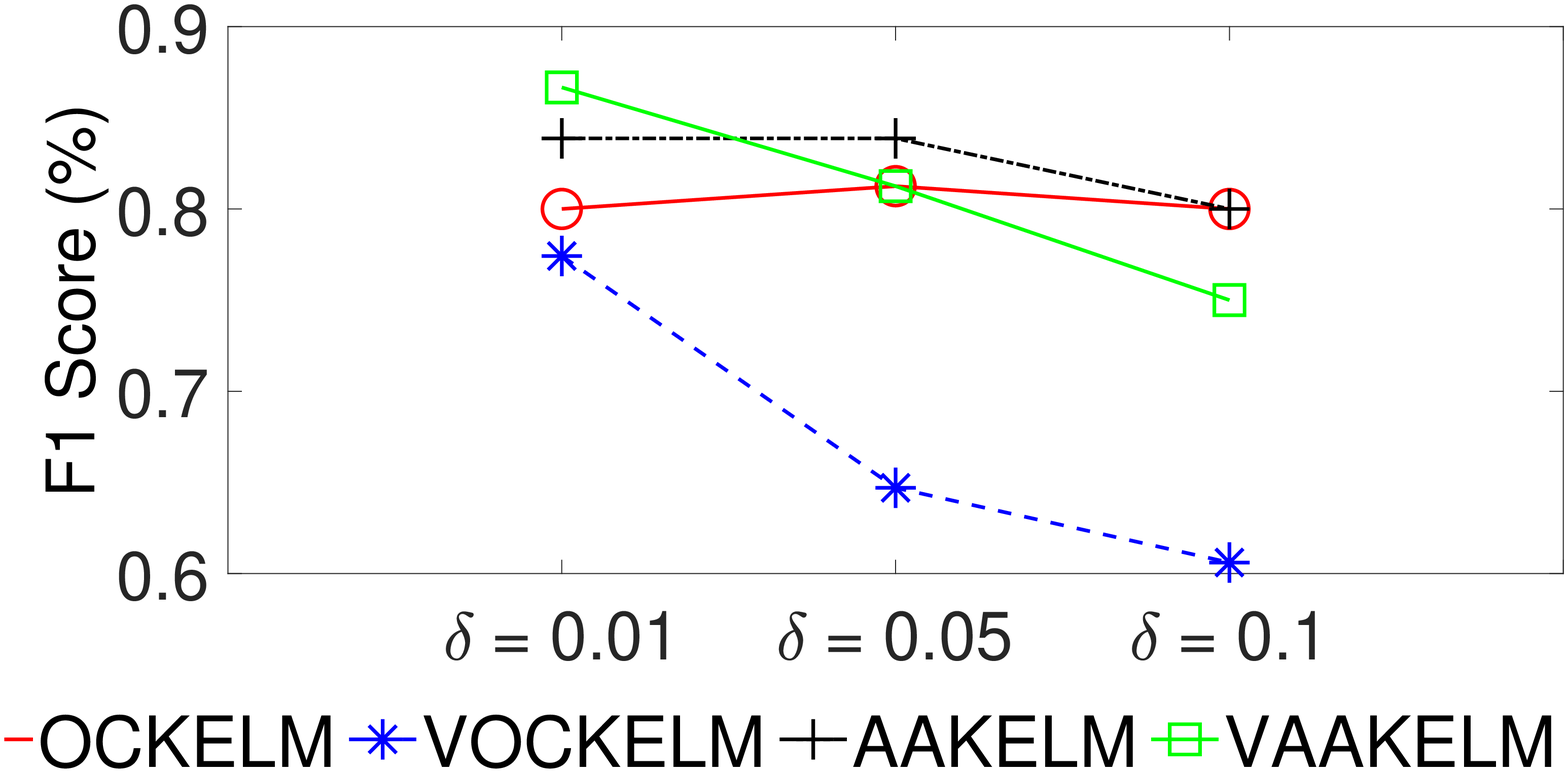}
			\caption{Wine}
		\end{subfigure}
		\caption{Variation of F$_1$ score with fraction of dismissal ($\delta$) for small-size one-class datasets.}
		\label{fig:mu_small}
	\end{center}
\end{figure*}

In OCC, the decision criteria is set during training time by taking a portion of data as outliers. We study the variation of F$_1$ score of different KELM-based classifiers across different values of fraction of dismissal, namely $\delta$ = 1\%, 5\%, 10\%, in Figure \ref{fig:mu_small}. The observations are noted as follows,
\begin{enumerate}
	\item For $\delta = 1\%$, \textit{VAAKELM} achieves the highest F$_1$ score for 12 datasets, displaying a clear advantage for Biomed, Cryotherapy, Diabetic Retinopathy, Imports, Vowel, and Waveform datasets.
	\item For $\delta = 5\%$, \textit{VAAKELM} achieves the highest F$_1$ score for 7 datasets, displaying a clear advantage for Breast Cancer, Diabetic Retinopathy, Imports, Vowel, and Waveform datasets.
	\item For $\delta = 10\%$, \textit{VAAKELM} achieves the highest F$_1$ score for 7 datasets, displaying a clear advantage for Diabetic Retinopathy, Heart Statlog, and Waveform datasets.
	\item For 5 datasets, \textit{VAAKELM} scores the highest across all $\delta$ values, while showing a clear advantage in performance for 2 datasets, namely, Diabetic Retinopathy, and Waveform.
\end{enumerate}
From the above observations, it can be inferred that \textit{VAAKELM} generally outperforms other classifiers across different values of $\delta$. Also, it can be observed that using small value of $\delta$ usually gives better results for small-size datasets.

\subsubsection{Experiments on medium-size datasets}\label{sec_exp_medium}

\begin{table*}[!b]
	\begin{center}
		\scriptsize
		\begin{tabular}{llll}
			\hline
			Target Class & \#Target & \#Outlier & \#Features \\
			\hline
			0            & 400    & 3600    & 256       \\
			1            & 400    & 3600    & 256       \\
			2            & 400    & 3600    & 256       \\
			3            & 400    & 3600    & 256       \\
			4            & 400    & 3600    & 256       \\
			5            & 400    & 3600    & 256       \\
			6            & 400    & 3600    & 256       \\
			7            & 400    & 3600    & 256       \\
			8            & 400    & 3600    & 256       \\
			9            & 400    & 3600    & 256       \\
			\hline
		\end{tabular}
		\caption{Specifications of concordia one-class datasets.}
		\label{tab:spec_medium}
		
		\hspace{2cm}
		
		\begin{tabular}{lllllll}
			\hline
			& \makecell{OCSVM\\\cite{scholkopf2001estimating}} & \makecell{SVDD\\\cite{tax2004support}}  & \makecell{OCKELM\\\cite{leng2015one}} & \makecell{VOCKLEM\\\cite{mygdalis2016one}} & \makecell{AAKELM\\\cite{gautam2017construction}} & VAAKLEM        \\
			\hline
			Class 0 & 75.39 & 80.7  & 54.88  & 65.56   & 84.29  & \textbf{85.11} \\
			Class 1 & 62.13 & 74.37 & 77.99  & 61.74   & \textbf{87.84}  & \textbf{87.84} \\
			Class 2 & 37.04 & 37.58 & 33.97  & 24.6    & 48.78  & \textbf{54.09} \\
			Class 3 & 43.18 & 47.89 & 46.32  & 43.8    & \textbf{70.87}  & \textbf{70.87} \\
			Class 4 & 48.32 & 51.13 & 32.21  & 27.52   & 58.25  & \textbf{59.41} \\
			Class 5 & 48.75 & 53.23 & 48.22  & 38.51   & 72.59  & \textbf{73.91} \\
			Class 6 & 73.2  & \textbf{76.47} & 43.38  & 35.08   & 71.8   & 73.2           \\
			Class 7 & 58.58 & 60    & 55.65  & 56.68   & 60.71  & \textbf{61.54} \\
			Class 8 & 34.3  & 39.43 & 32.36  & 31.15   & 48.8   & \textbf{51.52} \\
			Class 9 & 50.58 & 57.01 & 51.46  & 51.89   & 65.33  & \textbf{65.41} \\
			\hline
			$\eta_{\textbf{F}_1}$ & 53.15 & 57.78 & 47.64  & 43.65   & 66.93  & \textcolor{red}{\textbf{68.29}} \\
			\hline
		\end{tabular}
		\caption{F$_1$ score comparisons for different one-class classifiers for concordia one-class datasets.}
		\label{tab:specF1_medium}
	\end{center}
\end{table*}

\begin{figure*}[t]
	\begin{center}
		\begin{subfigure}{0.48\textwidth}
			\includegraphics[width=\linewidth]{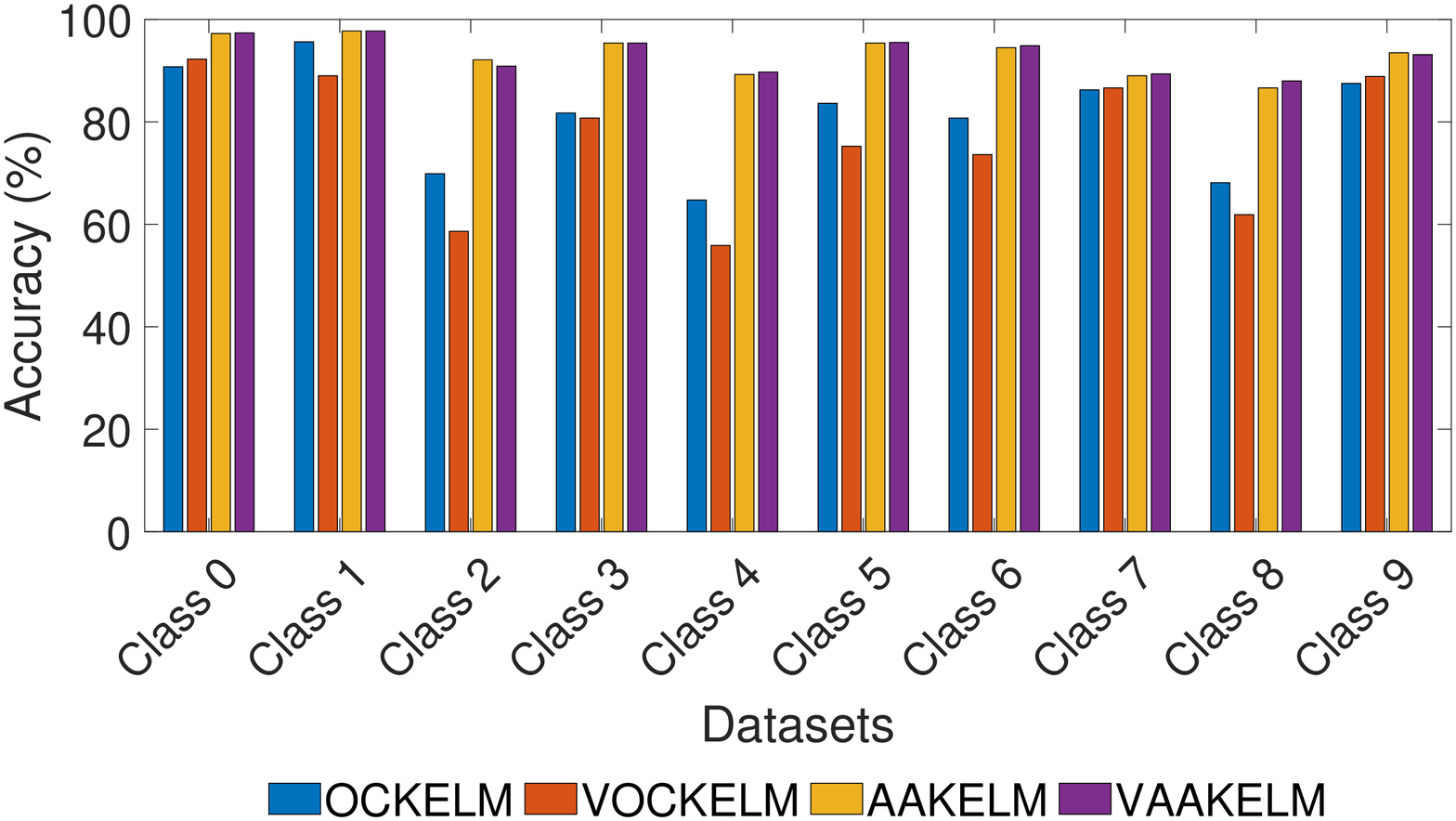}
			\caption{Accuracy}
		\end{subfigure}
		\begin{subfigure}{0.48\textwidth}
			\includegraphics[width=\linewidth]{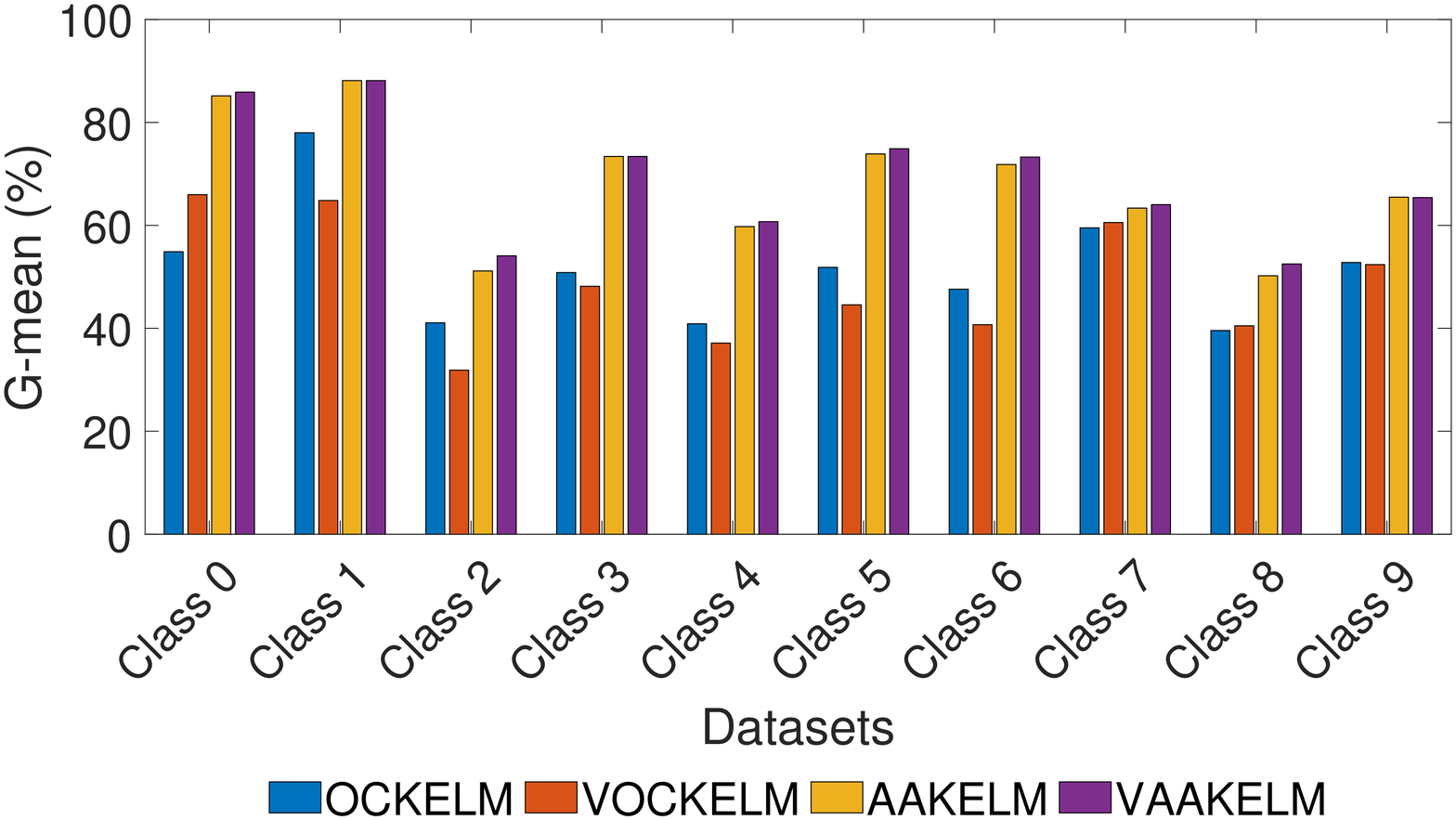}
			\caption{G-mean}
		\end{subfigure}
		
		\begin{subfigure}{0.48\textwidth}
			\includegraphics[width=\linewidth]{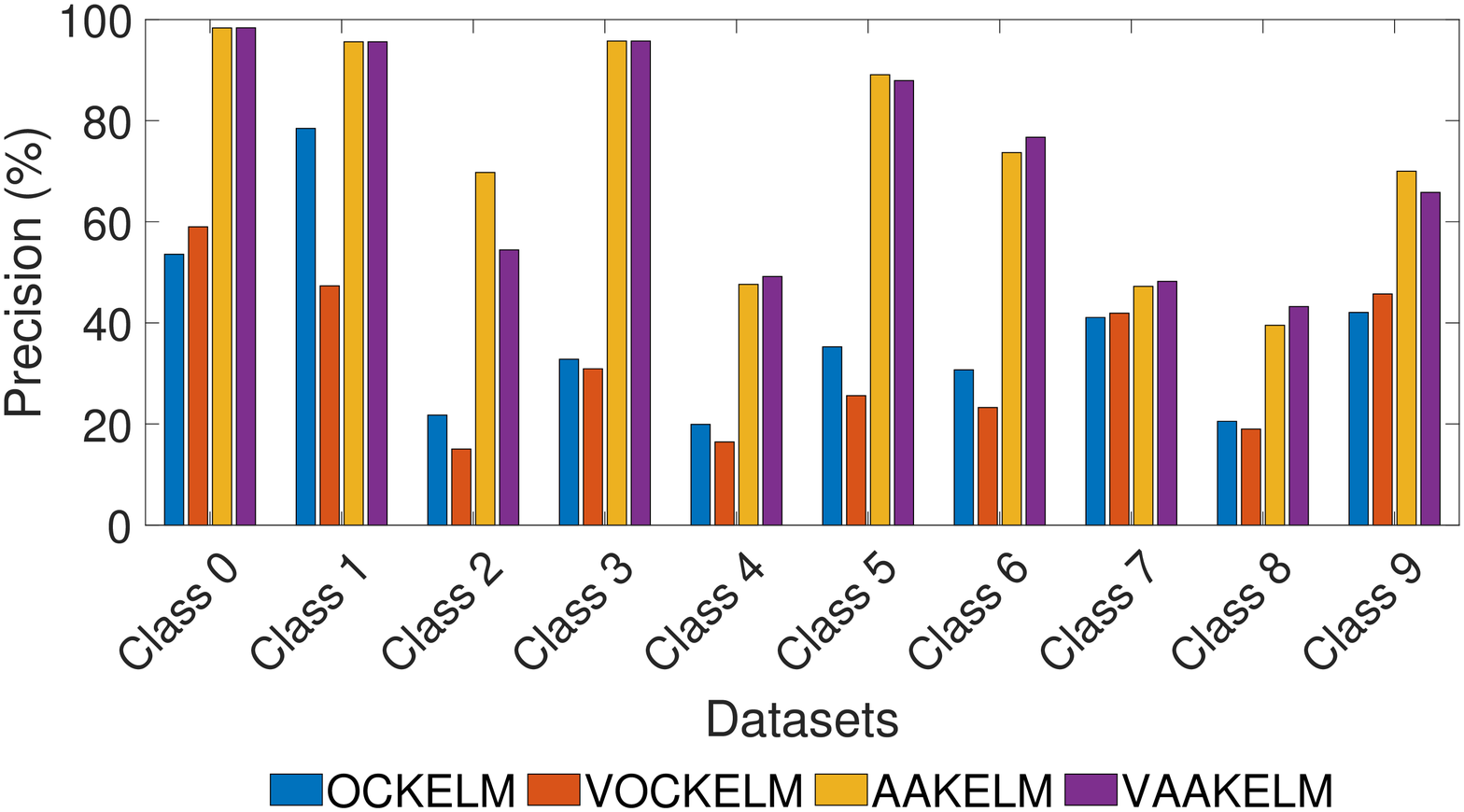}
			\caption{Precision}
		\end{subfigure}
		\begin{subfigure}{0.48\textwidth}
			\includegraphics[width=\linewidth]{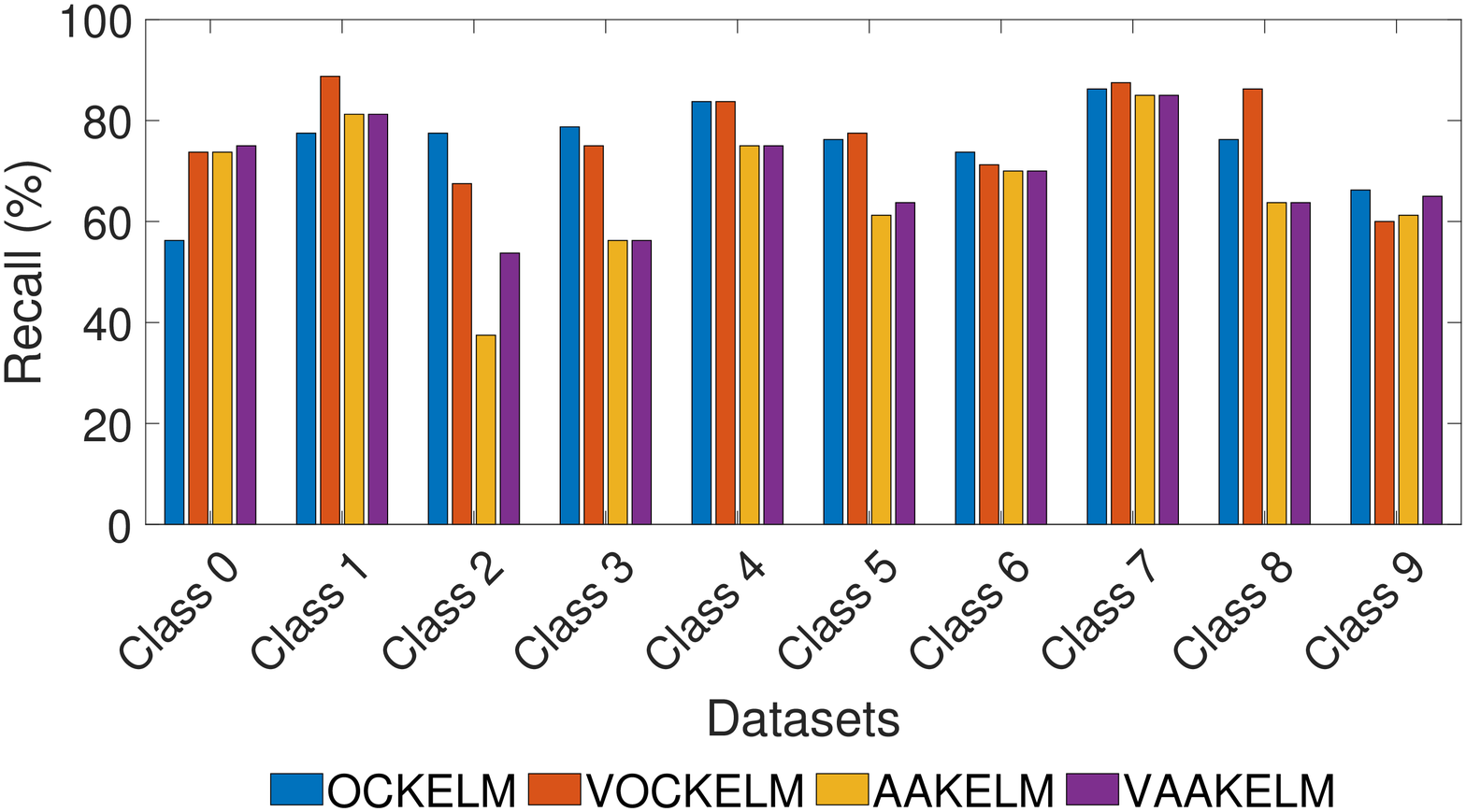}
			\caption{Recall}
		\end{subfigure}
		\caption{Examination of Accuracy, G-mean, Precision and Recall for concordia one-class datasets.}
		\label{fig:plotmetrics_medium}
	\end{center}
\end{figure*}

Further, we conduct experiments on 10 medium-size one-class datasets. These datasets are obtained from a single multi-class concordia digits dataset, by iteratively taking each class as the target class and rest of the classes as outliers. We provide the specifications of these datasets in Table \ref{tab:spec_medium}. We compare the performance of \textit{VAAKELM} with existing kernel-based one-class classifiers in Table \ref{tab:specF1_medium}, by considering $\eta_{\textbf{F}_1}$ as the final evaluation measure to rank the classifiers as per their performance. \textit{VAAKELM} obtains the highest $\eta_{\textbf{F}_1}$ (highlighted in bold red) as compared to the existing one-class classifiers, with a significant improvement of 10.51\% in comparison to non-KELM-based classifiers (i.e., \textit{OCSVM} and \textit{SVDD}). Further, \textit{VAAKELM} showed an improvement of 20.65\% in comparison to boundary-based KELM classifiers (i.e., \textit{OCKELM} and \textit{VOCKELM}), in terms of $\eta_{\textbf{F}_1}$. It can be noted that \textit{VAAKELM} obtains the highest F$_1$ score for 9 out of 10 datasets (highlighted in bold). From the above observations, it can be concurred that \textit{VAAKELM} can outperform the existing kernel-based one-class classifiers for medium-size datasets, and can act as a suitable alternative for OCC tasks. For reference, we also present the experimental results based on accuracy,
g-mean, precision, and recall metrics in Figure \ref{fig:plotmetrics_medium}. Out of 10 datasets, \textit{VAAKELM} scores the highest accuracy, and g-mean for an overwhelming 8, 9 datasets, respectively. Also, it scores the highest precision for 7 datasets.

\begin{figure*}[t]
	\begin{center}
		\begin{subfigure}{0.3\textwidth}
			\includegraphics[width=\linewidth]{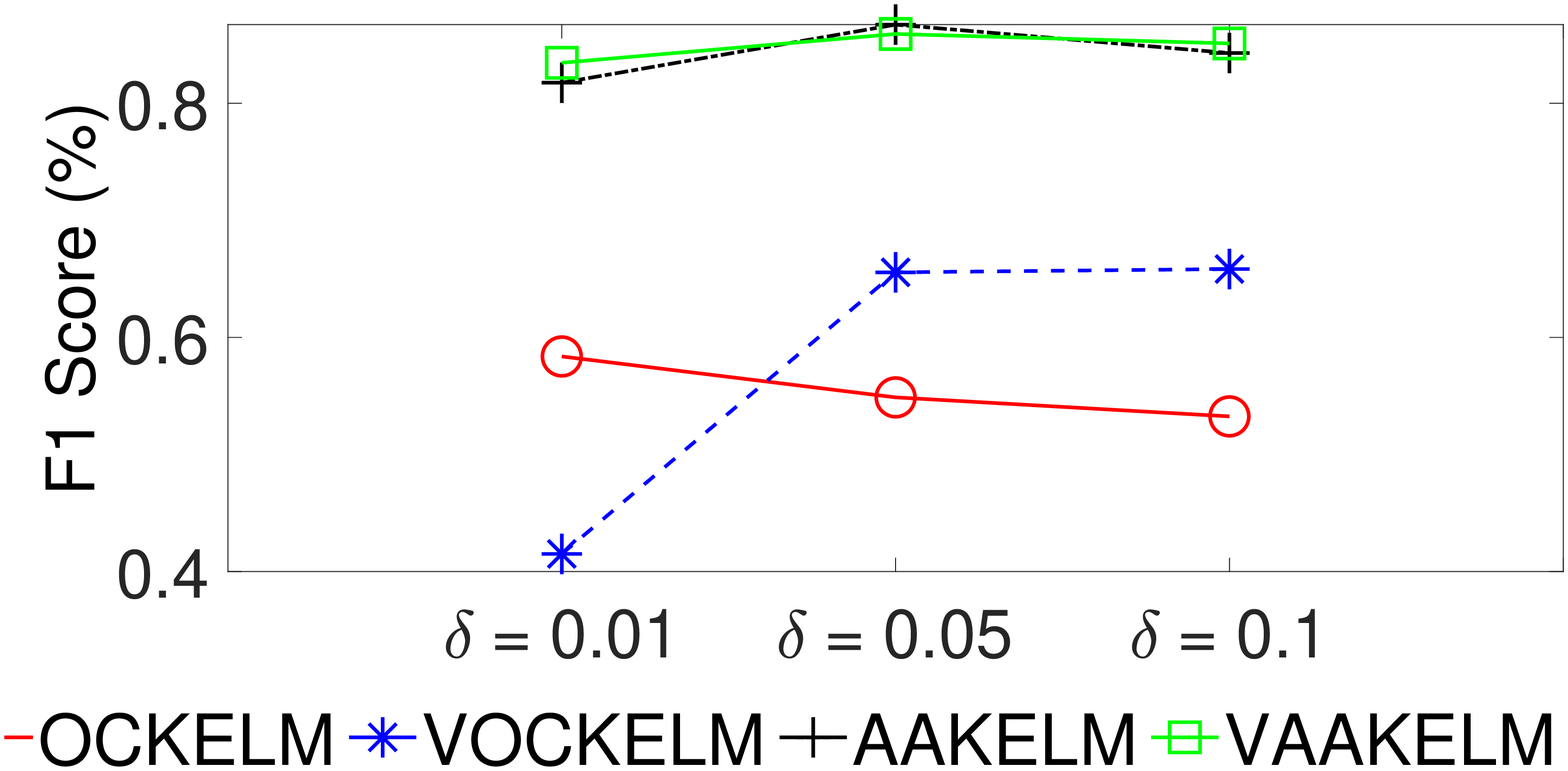}
			\caption{Class 0}
		\end{subfigure}
		\begin{subfigure}{0.3\textwidth}
			\includegraphics[width=\linewidth]{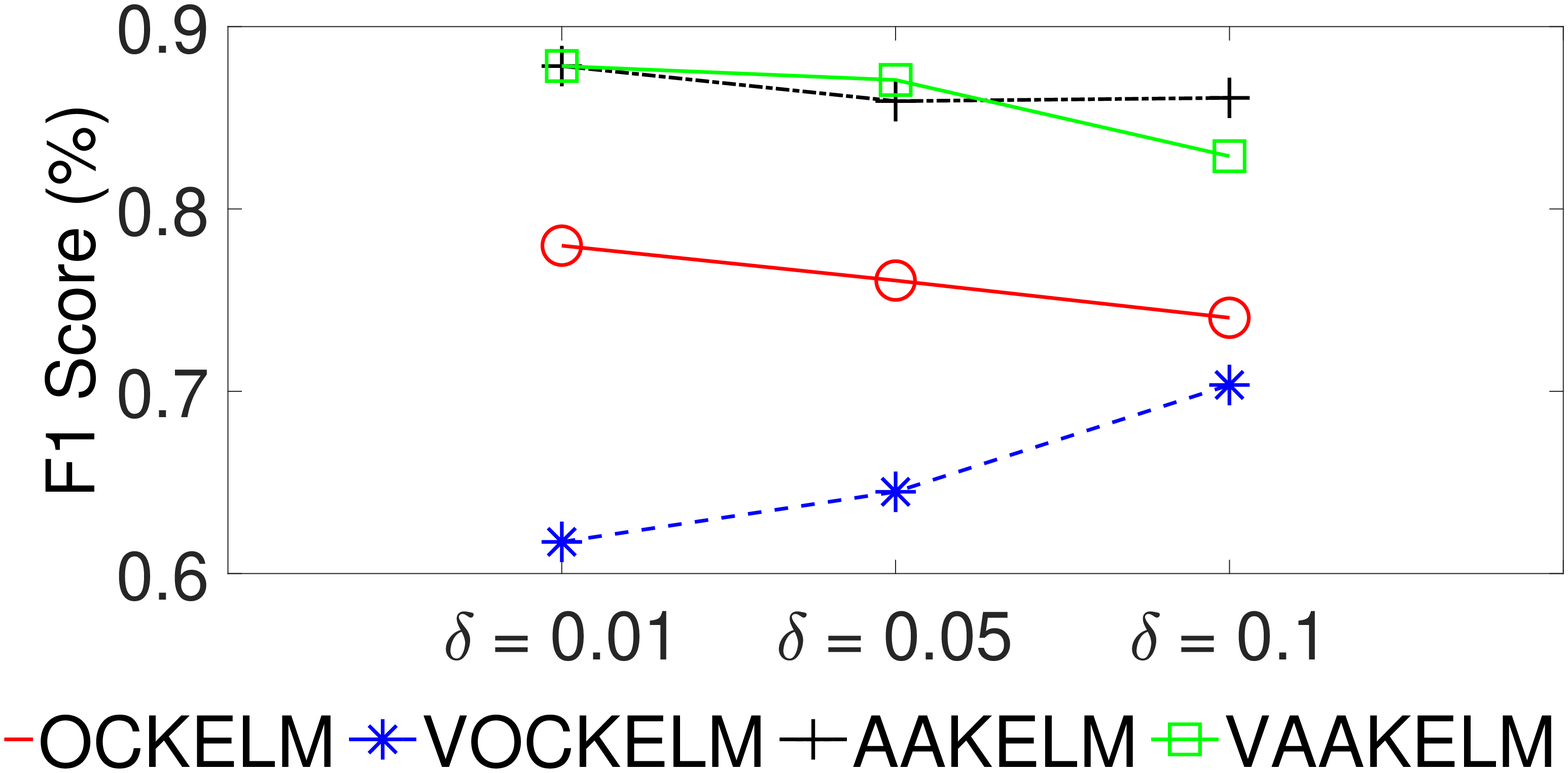}
			\caption{Class 1}
		\end{subfigure}
		\begin{subfigure}{0.3\textwidth}
			\includegraphics[width=\linewidth]{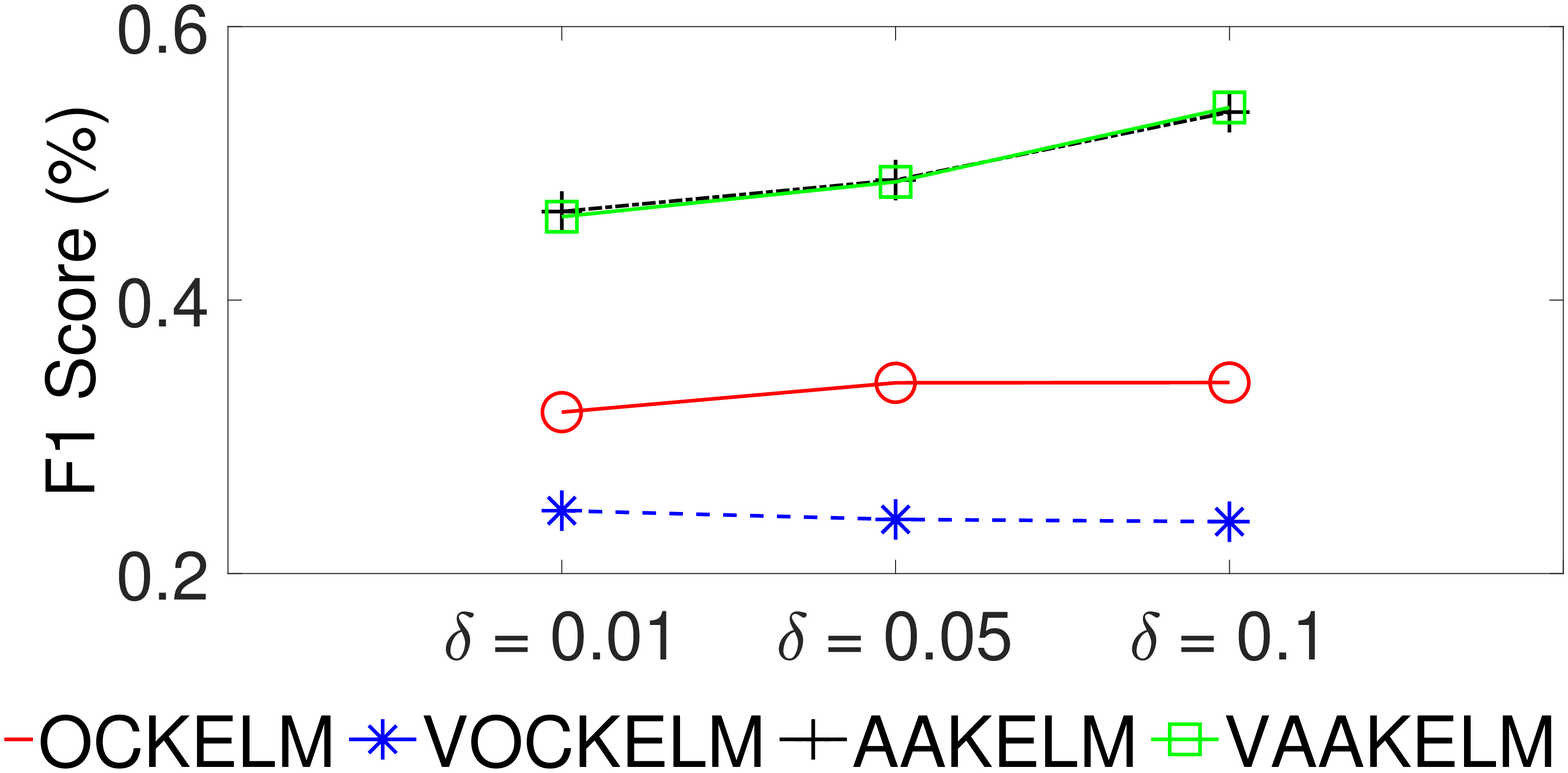}
			\caption{Class 2}
		\end{subfigure}
		
		\begin{subfigure}{0.3\textwidth}
			\includegraphics[width=\linewidth]{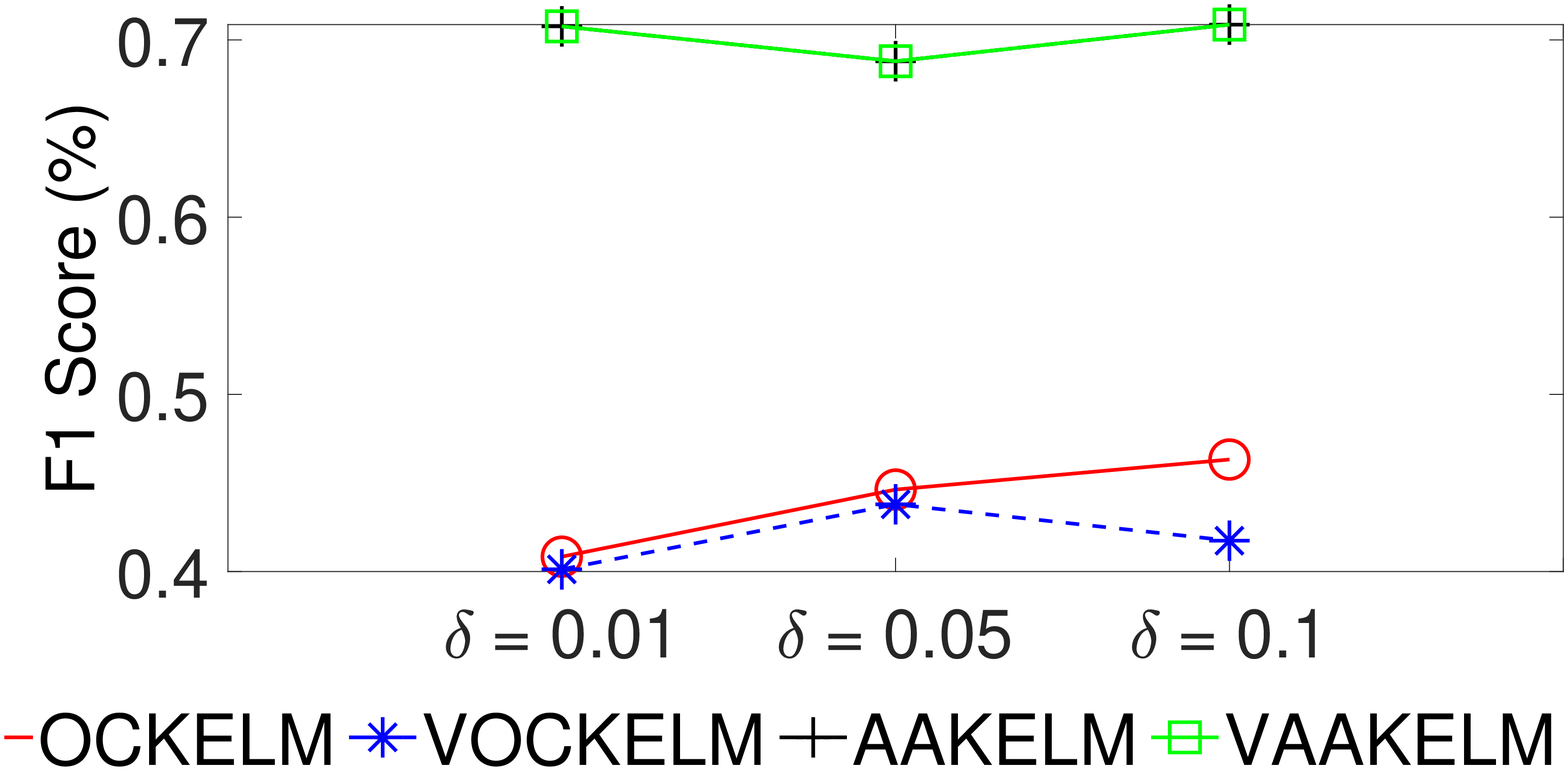}
			\caption{Class 3}
		\end{subfigure}
		\begin{subfigure}{0.3\textwidth}
			\includegraphics[width=\linewidth]{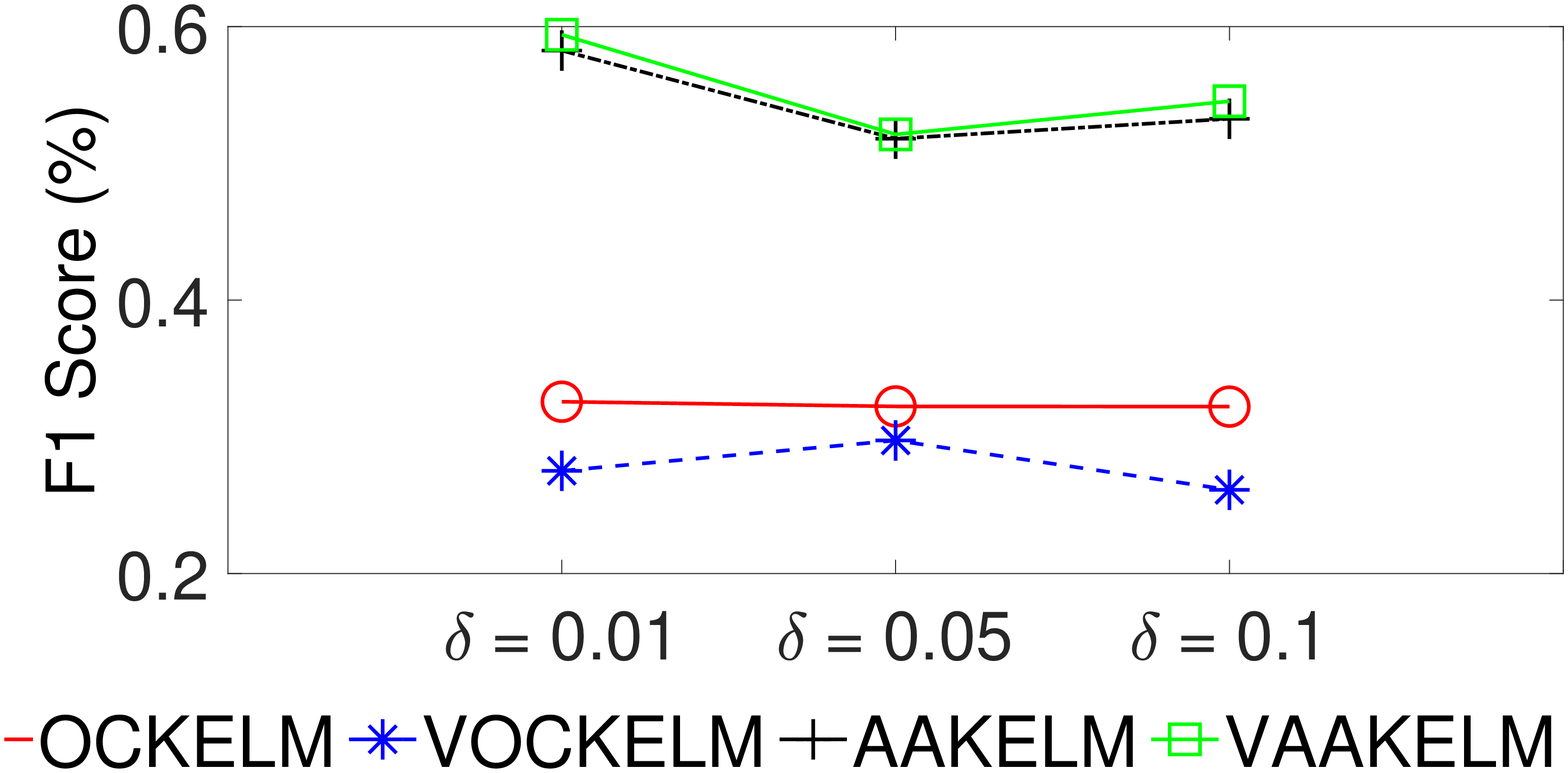}
			\caption{Class 4}
		\end{subfigure}
		\begin{subfigure}{0.3\textwidth}
			\includegraphics[width=\linewidth]{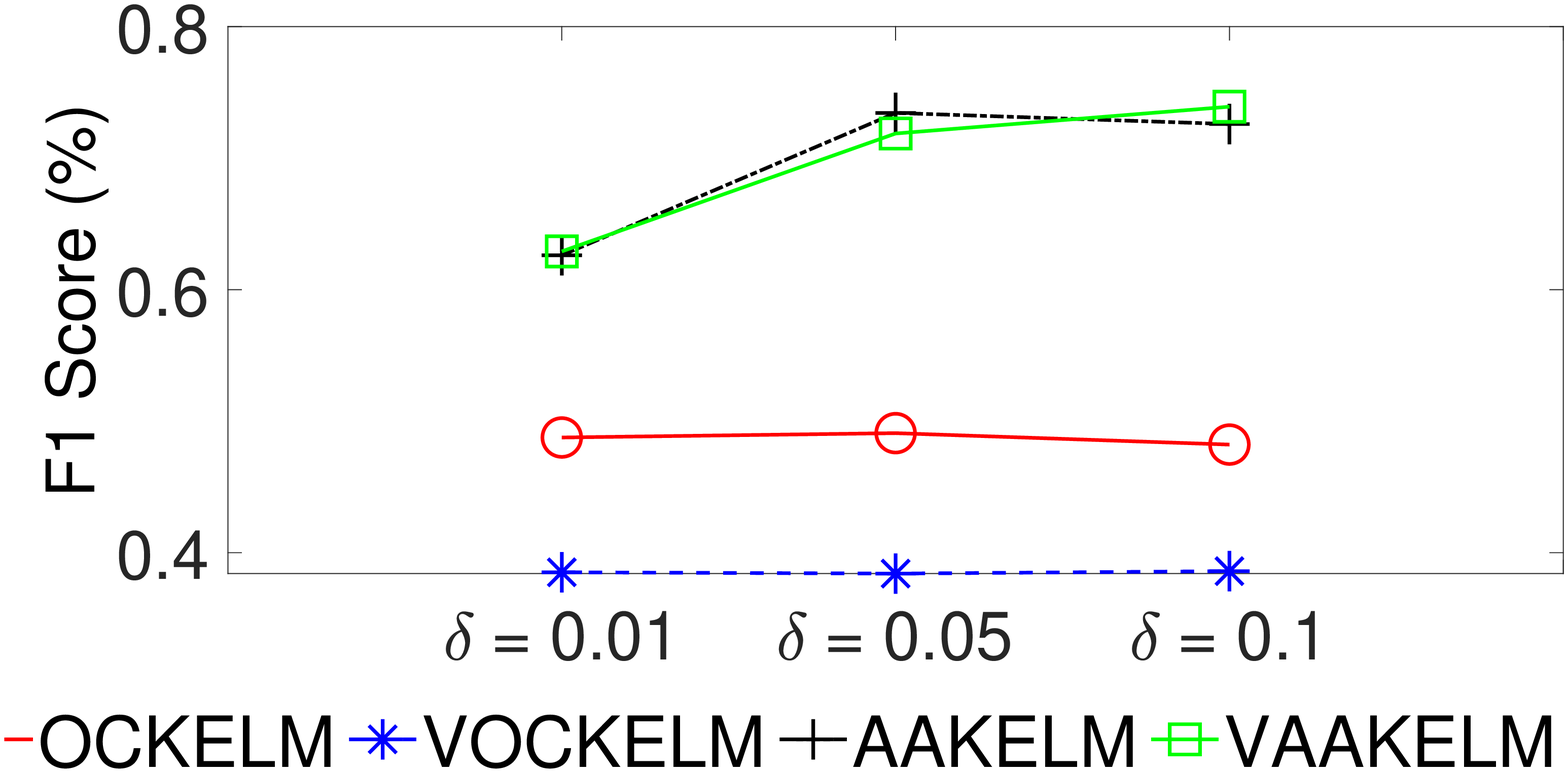}
			\caption{Class 5}
		\end{subfigure}
		
		\begin{subfigure}{0.3\textwidth}
			\includegraphics[width=\linewidth]{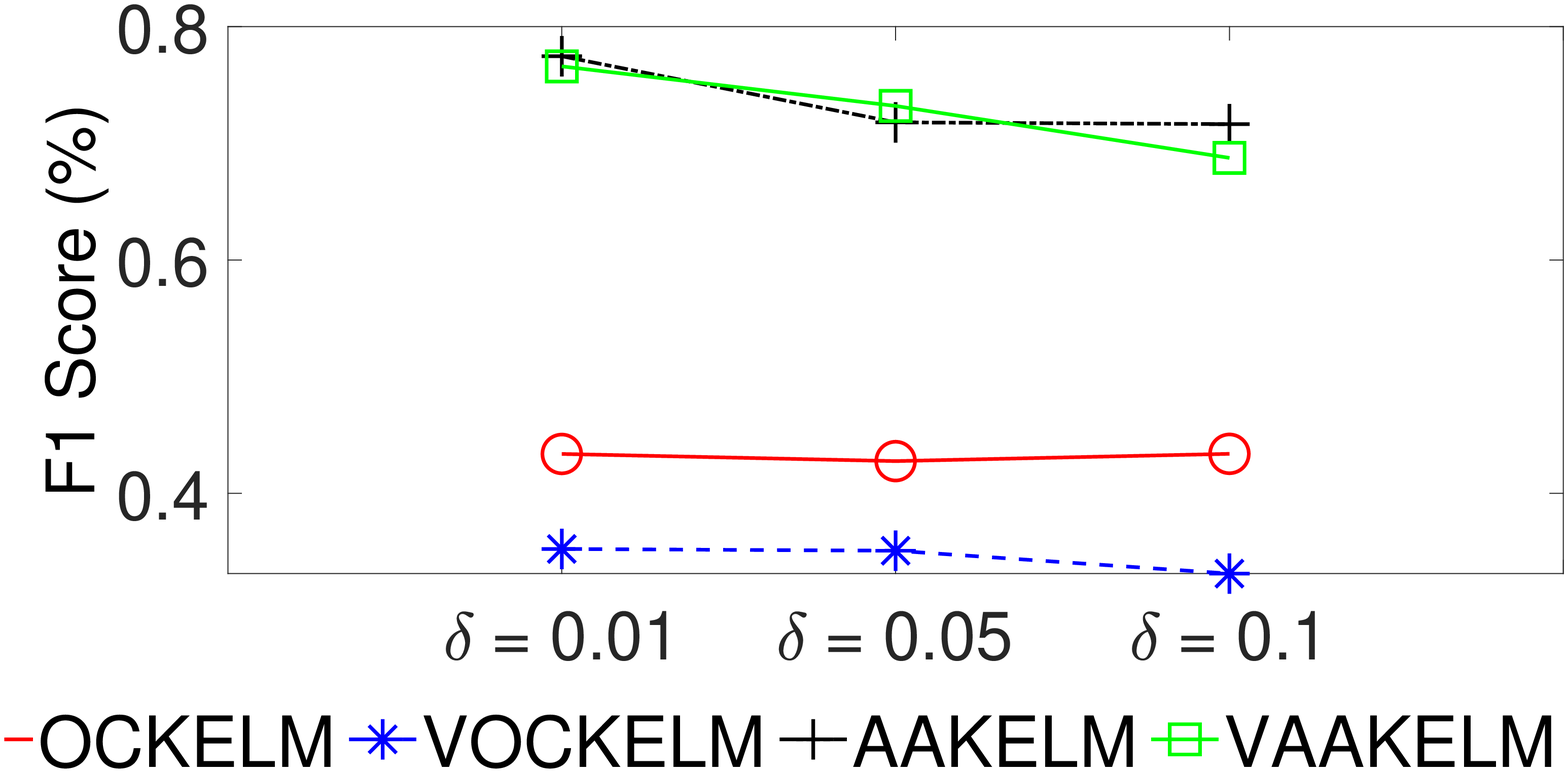}
			\caption{Class 6}
		\end{subfigure}
		\begin{subfigure}{0.3\textwidth}
			\includegraphics[width=\linewidth]{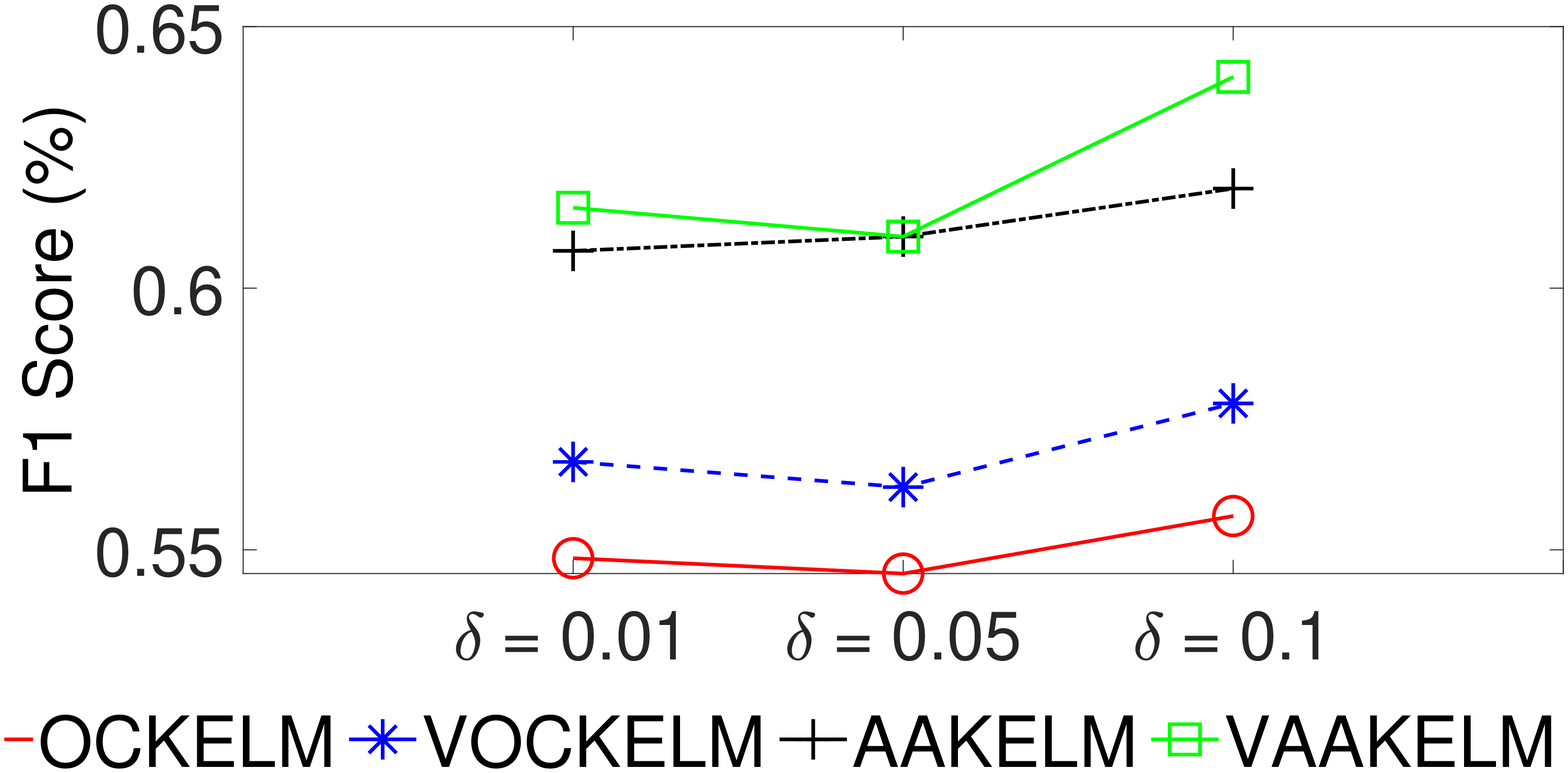}
			\caption{Class 7}
		\end{subfigure}
		\begin{subfigure}{0.3\textwidth}
			\includegraphics[width=\linewidth]{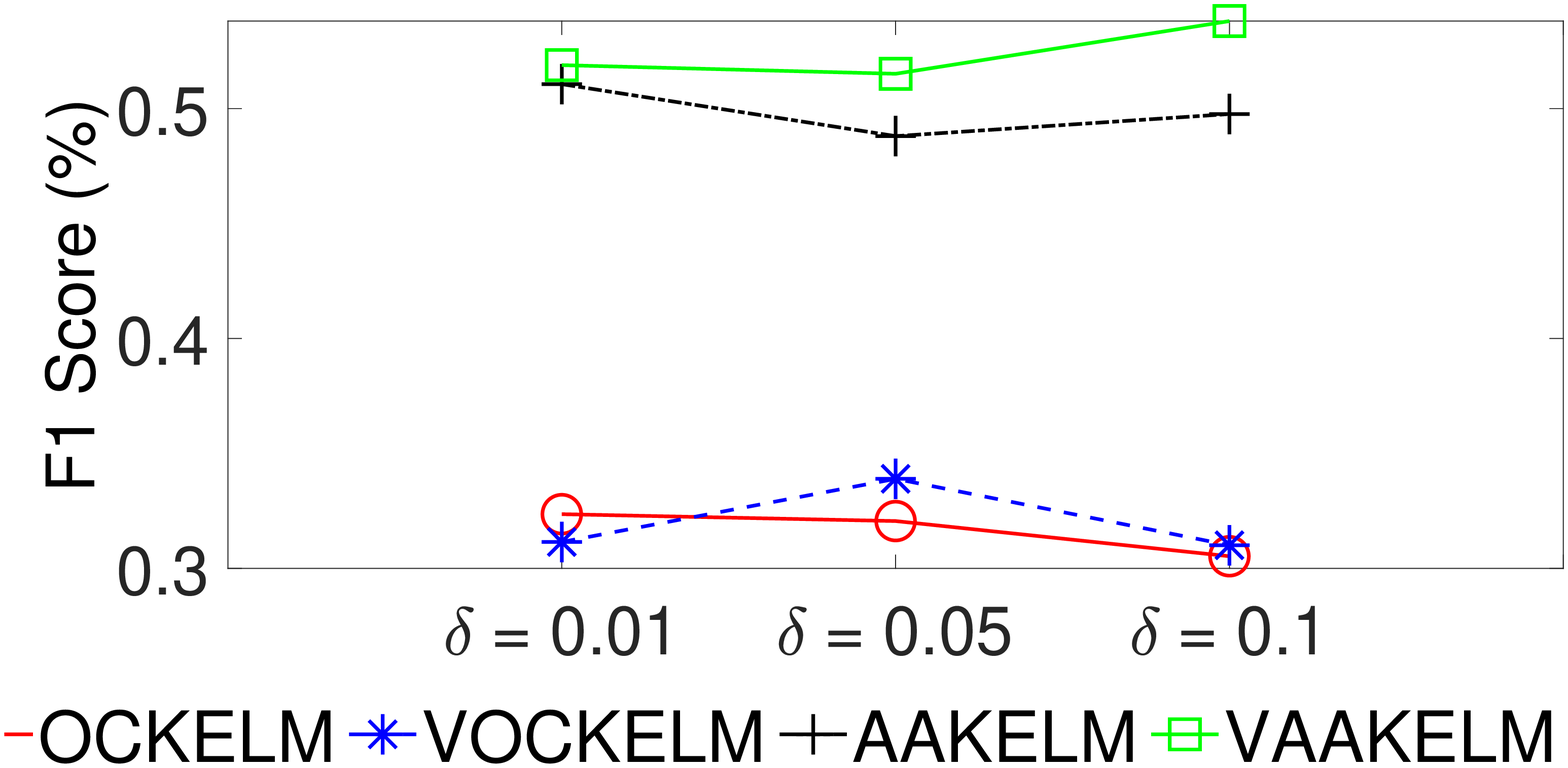}
			\caption{Class 8}
			\label{concord_class8}
		\end{subfigure}
		
		\begin{subfigure}{0.3\textwidth}
			\includegraphics[width=\linewidth]{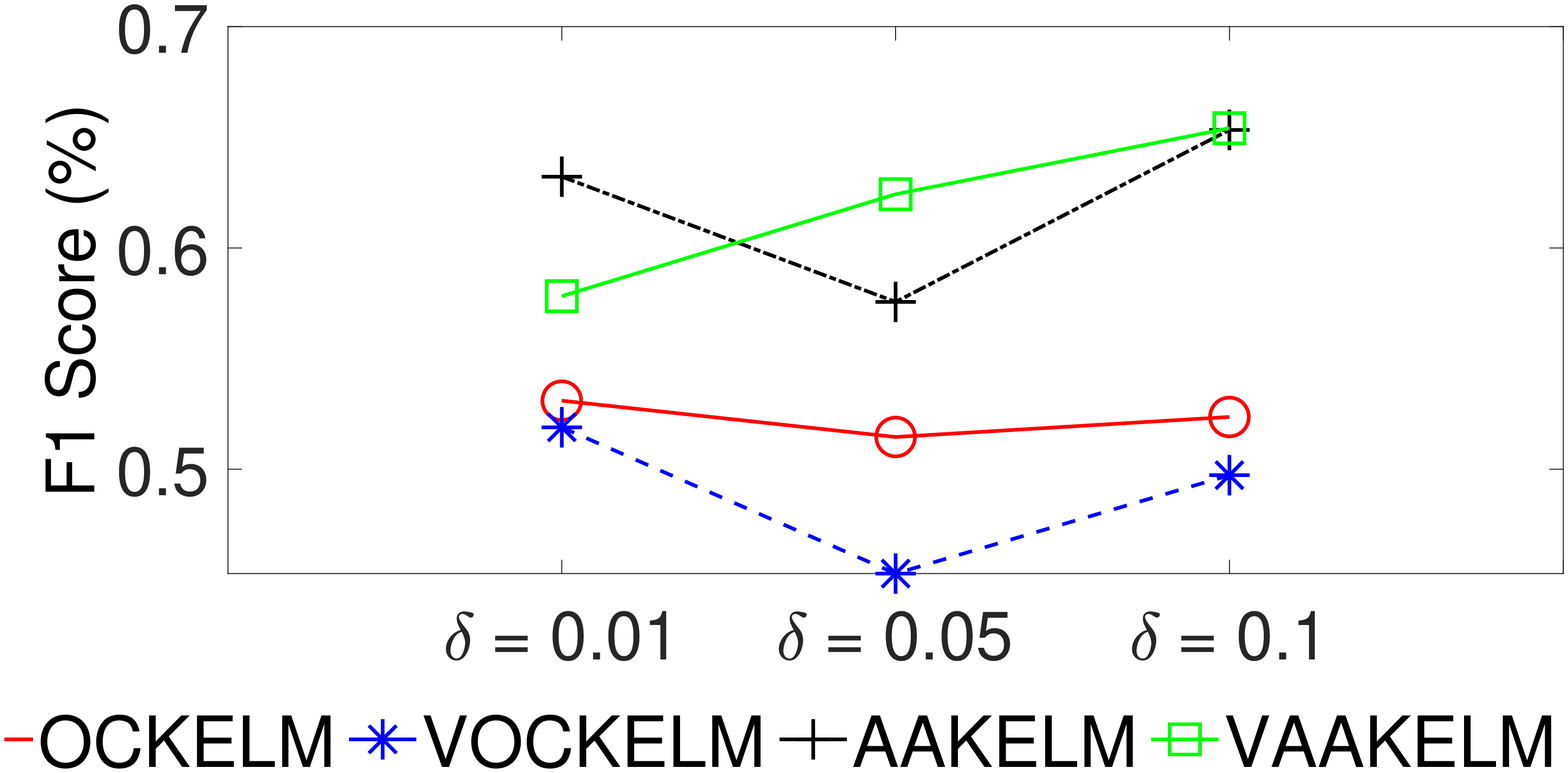}
			\caption{Class 9}
		\end{subfigure}
		\caption{Variation of F$_1$ score with fraction of dismissal ($\delta$) for concordia one-class datasets.}
		\label{fig:mu_medium}
	\end{center}
\end{figure*}

Similarly, we study the variation of F$_1$ score for different KELM-based one-class classifiers across different values of fraction of dismissal, namely $\delta$ = 1\%, 5\%, 10\%, for medium-size one-class datasets in Figure \ref{fig:mu_medium}. For $\delta = 1\%$, and $5\%$, \textit{VAAKELM} achieves the highest F$_1$ score for 8 datasets as compared to other one-class classifiers. Also, \textit{VAAKELM} shows a significant advantage over other classifiers for the case of class 8 in Figure \ref{concord_class8}. Further, it can be noted that for all 10 cases in Figure \ref{fig:mu_medium}, the reconstruction-based classifiers (i.e., \textit{AAKELM} and \textit{VAAKELM}) show a significant advantage over boundary-based classifiers (i.e., \textit{OCKELM} and \textit{VOCKELM}).

\begin{table*}[h]
	\centering
	\fontsize{5}{8}\selectfont
	\begin{tabular}{lllllllllll}
		\hline
		& \makecell{OCRF\\\cite{desir2013one}}   & \makecell{Naive Parzen\\\cite{duin1976choice}} & \makecell{k-means\\\cite{jiang2001two}} & \makecell{k-NN\\\cite{knorr2000distance}}   & \makecell{Autoencoder\\\cite{tax2002one}} & \makecell{PCA\\\cite{bishop1995neural}}    & \makecell{MST\\\cite{juszczak2009minimum}}    & \makecell{kCentre\\\cite{hochbaum1985best}} & \makecell{SVDD\\\cite{tax2004support}}   & VAAKELM         \\
		\hline
		Biomed                 & 0.1541 & 1.7659       & 0.0835  & 0.025  & 1.9163      & 0.1461 & 0.0204 & 0.4048    & 0.1421 & \textbf{0.0175} \\
		Breast Cancer          & 0.3636 & 0.7067       & 0.0203  & 0.027  & 0.766       & 0.0751 & 0.0223 & 0.3626    & 0.2242 & 0.0323          \\
		Cardiotocography       & 0.489  & 1.5114       & 0.0199  & 0.0203 & 0.8285      & 0.0644 & 0.0224 & 0.3076    & 0.0484 & 0.0536          \\
		Colposcopy             & 1.8691 & 0.7467       & 0.0195  & 0.0253 & 4.3599      & 0.1073 & 0.021  & 0.5853    & 0.0287 & \textbf{0.0129} \\
		Cryotherapy            & 0.072  & 0.0877       & 0.0202  & 0.0141 & 2.1011      & 0.0818 & 0.0135 & 0.3157    & 0.0691 & \textbf{0.0110} \\
		Diabetic   Retinopathy & 1.4402 & 1.8035       & 0.0268  & 0.0358 & 3.8014      & 0.0391 & 0.0261 & 0.4515    & 0.1447 & 0.1173          \\
		Ecoli                  & 0.076  & 0.0928       & 0.0152  & 0.0136 & 0.2588      & 0.0391 & 0.0134 & 0.2986    & 0.0147 & \textbf{0.0032} \\
		Heart   Cleveland      & 0.2963 & 0.6452       & 0.0156  & 0.0159 & 2.3313      & 0.0344 & 0.0177 & 0.2914    & 0.0229 & \textbf{0.0147} \\
		Heart Statlog          & 0.217  & 0.1841       & 0.0149  & 0.0202 & 0.5304      & 0.0348 & 0.0146 & 0.3006    & 0.0131 & \textbf{0.0077} \\
		Imports                & 0.749  & 0.3033       & 0.0147  & 0.0158 & 0.7712      & 0.0459 & 0.0162 & 0.2921    & 0.0263 & \textbf{0.0051} \\
		Sonar                  & 2.4491 & 0.7062       & 0.0153  & 0.0134 & 10.7223     & 0.0354 & 0.0155 & 0.2981    & 0.0135 & \textbf{0.0055} \\
		Survival               & 0.2179 & 0.3324       & 0.0155  & 0.018  & 0.3082      & 0.0358 & 0.0199 & 0.3413    & 0.1616 & 0.0185          \\
		Vowel                  & 0.0921 & 0.1335       & 0.0137  & 0.0228 & 0.2363      & 0.0353 & 0.0145 & 0.3054    & 0.0103 & \textbf{0.0041} \\
		Waveform               & 0.8212 & 2.3675       & 0.0149  & 0.023  & 0.6854      & 0.0348 & 0.0249 & 0.3286    & 0.0379 & 0.0491          \\
		Wine                   & 0.138  & 0.1654       & 0.0168  & 0.0162 & 0.1591      & 0.0428 & 0.0159 & 0.2873    & 0.0152 & \textbf{0.0059} \\
		\hline
	\end{tabular}
	\caption{Training time (in secs) for different classifiers.}
	\label{tab:trtime}
\end{table*}

The KELM-based one-class classifiers follow a non-iterative approach to learning, hence take less training time. We present the training time spent on \textit{VAAKELM} and the existing non-KELM-based one-class classifiers in Table \ref{tab:trtime}. The training time of \textit{OCSVM} is not recorded in the table as \textit{OCSVM} used the Mex C++ compiler and not the same environment as other classifiers (i.e., MATLAB). It can be observed that the training time for \textit{VAAKELM} is mostly least (highlighted in bold in Table \ref{tab:trtime}) in comparison to the other one-class classifiers. The lesser time taken by \textit{VAAKELM} in comparison to the other classifiers can be accredited to its non-iterative nature.

\section{Conclusion} \label{sec_conclusion}
In this paper, we proposed the minimum variance embedded auto-associative kernel extreme learning machine for OCC (\textit{VAAKELM}). It is a single-layer method and follows a reconstruction-based approach to OCC. The minimum variance embedding reduces the variance of the target class data and forces the network output weights to emphasize in regions of low variance. \textit{VAAKELM} uses reconstruction error to define a threshold criterion to decide the membership of a data sample. The KELM-based autoencoder helps to learn the essential information of the input data at the output layer. We experimented with \textit{VAAKELM} on 15 small-size and 10 medium-size one-class datasets and compared its performance with 13 existing one-class classifiers. For small-size datasets, \textit{VAAKELM} obtained the highest $\eta_{\textbf{F}_1}$ as compared to the existing one-class classifiers, with a significant improvement of 7.47\% in comparison to non-kernel-based classifiers and 4.53\% in comparison to other kernel-based classifiers. For medium-size datasets, \textit{VAAKELM} obtained the highest $\eta_{\textbf{F}_1}$ with a significant improvement of 10.51\% in comparison to non-KELM-based classifiers (i.e., \textit{OCSVM} and \textit{SVDD}). Further, \textit{VAAKELM} showed an improvement of 20.65\% in comparison to boundary-based KELM classifiers (i.e., \textit{OCKELM} and \textit{VOCKELM}). It can be concurred that \textit{VAAKELM} can outperform the existing one-class classifiers, and can act as a suitable alternative to the existing one-class classifiers.

\textit{VAAKELM} involves the computation of the inverse of the output weight matrix, with a complexity of $O(n^3)$. With an increase in the size of training data, the computational overhead of the inverse matrix operation increases drastically. Further research can be done to tackle this problem and make \textit{VAAKELM} scalable for big data handling. We developed \textit{VAAKELM} for an offline setting to handle stationary data. It can be further extended for online learning to handle non-stationary data.


%
%

\section*{Conflict of interest}
The authors declare that they have no conflict of interest.

\bibliographystyle{spbasic}      
\bibliography{ms}


\end{document}